\documentclass[10pt,twocolumn,letterpaper]{article}

\usepackage[pagenumbers]{cvpr}             

\usepackage[table]{xcolor}

\definecolor{mylightred}{RGB}{253, 231, 231}
\definecolor{mylightyellow}{RGB}{255, 255, 224}
\definecolor{mylightgreen}{RGB}{229, 245, 229}
\definecolor{mylightblue}{RGB}{227, 235, 255}

\definecolor{cvprblue}{rgb}{0.21,0.49,0.74}
\usepackage[pagebackref,breaklinks,colorlinks,allcolors=cvprblue]{hyperref}

\usepackage{subcaption}
\usepackage{blindtext}
\usepackage{graphicx}
\usepackage{bm}
\usepackage{colortbl}
\usepackage{multirow}
\usepackage[table]{xcolor} 
\usepackage{multirow}
\usepackage{booktabs}
\usepackage{makecell}  
\usepackage{array} 
\newcolumntype{L}[1]{>{\raggedright\arraybackslash}p{#1}}
\usepackage{marvosym} 

\renewcommand{\paragraph}[1]{{\vspace{0.01in}\noindent\textbf{#1}}~}
\definecolor{color_ours}{gray}{0.9}

\newcommand{\modelname}{{Mirai}}

\newcommand{\modelunidir}{\mbox{Mirai-E}}
\newcommand{\modelbidir}{\mbox{Mirai-I}}

\newcommand{\x}{\bm{x}}

\newcommand{\X}{\mathbf{X}}

\def\paperID{} 
\def\confName{CVPR}
\def\confYear{2026}

\title{Mirai: Autoregressive Visual Generation Needs Foresight}

\author{
Yonghao Yu$^{1}$ \quad
Lang Huang$^{2}$ \footnotemark[1]  \quad
Zerun Wang$^{1}$ \quad
Runyi Li$^{3}$ \quad
Toshihiko Yamasaki$^{1}$\vspace{0.3em} \\
$^{1}$The University of Tokyo \quad
$^{2}$National Institute of Informatics \quad
$^{3}$Peking University  \quad \\
{\tt\small \{y\_yu, ze\_wang, yamasaki\}@cvm.t.u-tokyo.ac.jp,} {\tt\small lang@nii.ac.jp, lirunyi@stu.pku.edu.cn}
}

\begin{document}

\twocolumn[{%
\renewcommand\twocolumn[1][]{#1}%
\maketitle
\begin{center}
\vspace{-0.38in}
    \centering
    \captionsetup{type=figure} 
    
    \begin{subfigure}{0.58\textwidth} 
        \centering
        \includegraphics[width=\textwidth]{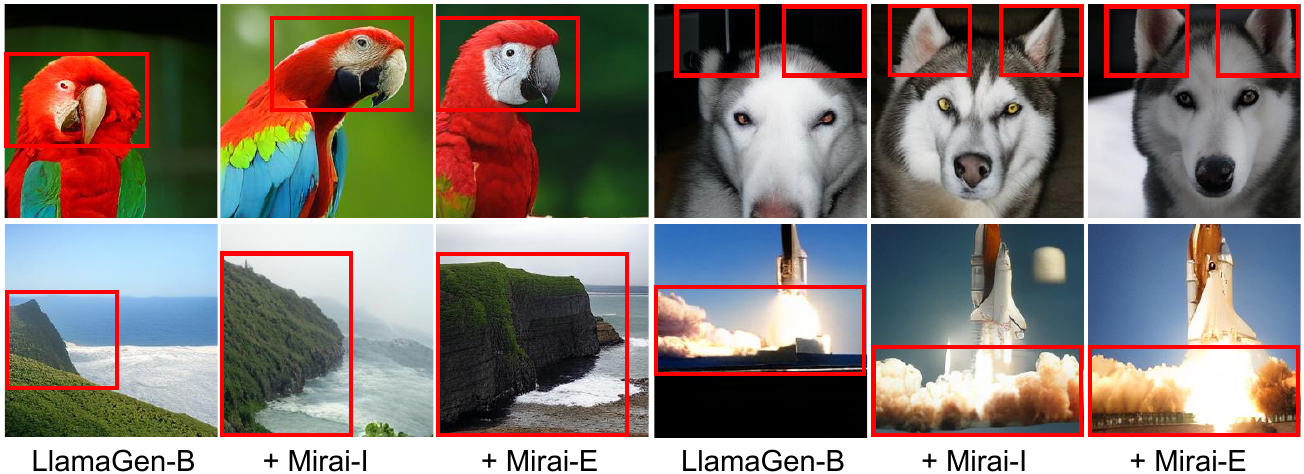} 
        \vspace{-3.5mm}
        \label{fig:overviewa}
    \end{subfigure}
    \hspace{2em}
    \begin{subfigure}{0.29\textwidth}
        \centering
        \includegraphics[width=\textwidth]{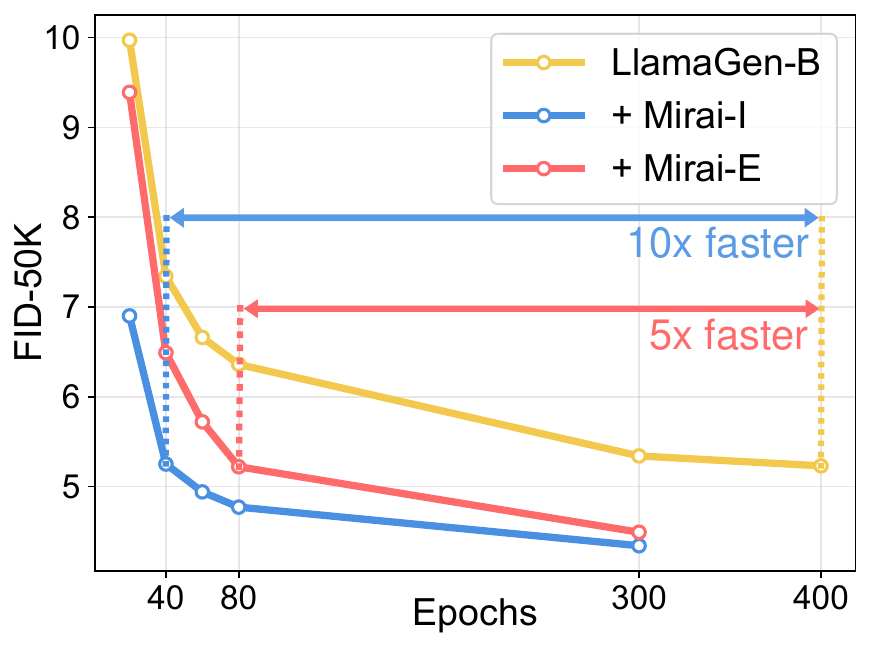} 

        \label{fig:overviewb}
    \end{subfigure}
    
    \setcounter{figure}{0}
       \vspace{-3.5mm}
    \captionof{figure}{\emph{Left}: The sample comparison between the AR baseline LlamaGen-B \cite{sun2024autoregressive} and our \modelname{} with the same 300-epoch training. The area enclosed by the red rectangle demonstrates the global consistency of images generated by our method. For example, in the rocket launch scene (bottom row right), the baseline model fails to maintain global structure, rendering a misaligned smoke. In contrast, our method generates a complete and structurally coherent result.
     \emph{Right}: The performance of our \modelname{} on training acceleration. We quantify this effect across multiple samples in Sec.~\ref{sec:system-level}} 
    \label{fig:overview} 
    \vspace{-2.5mm}
\end{center}
}]

\footnotetext[1]{Corresponding author.}
\begin{abstract}
Autoregressive (AR) visual generators model images as sequences of discrete tokens and are trained with a next-token likelihood objective. This strict causal supervision optimizes each step based only on the immediate next token, which can weaken global coherence and slow convergence. We investigate whether \textbf{foresight}, training signals that originate from later tokens, can improve autoregressive visual generation. We conduct a series of controlled diagnostics along the injection level, foresight layout, and foresight source axes, revealing a key insight: aligning foresight with AR models' internal representations on the 2D image grid improves causal modeling.
We formulate this insight with \modelname{} (meaning ``future'' in Japanese), a general framework that injects future information into AR training with no architecture change and no extra inference overhead: 
\modelunidir{} uses explicit foresight from multiple future positions of unidirectional representations, whereas \modelbidir{} leverages implicit foresight from matched bidirectional representations.
Extensive experiments show that \modelname{} significantly accelerates convergence and improves generation quality. For instance, \modelbidir{} speeds up LlamaGen-B's convergence by up to 10$\times$ and reduces the generation FID from 5.34 to 4.34 on the ImageNet class-condition image generation benchmark.
Our study highlights that visual autoregressive models need foresight. Project Page: \url{https://y0uroy.github.io/Mirai}.
\end{abstract}

\section{Introduction}
\label{sec:intro}
\begin{figure*}[t]
  \vspace{-0.1in}
  \centering
    \includegraphics[width=1\linewidth]{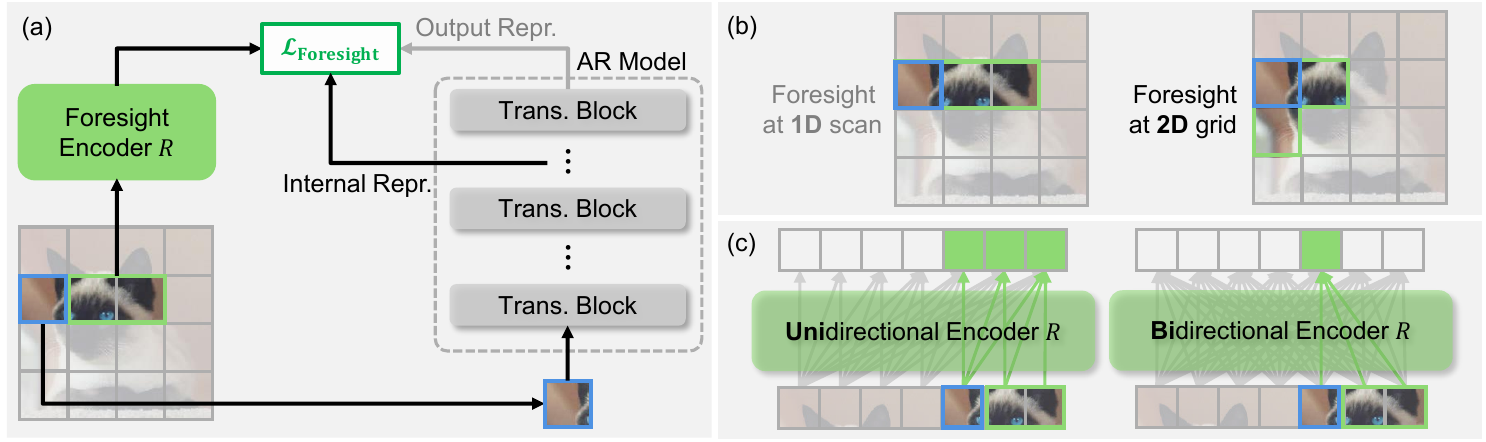}
     \vspace{-6mm}
    \caption{\textbf{Overview of our explorations in the visual AR with foresight}. For illustration, all subfigures (except the right subfigure of (c)) use $K = 3$ foresight tokens here. (a) Foresight injection level. (b) Foresight in 1D scan \emph{vs.}\ 2D grid. (c) The source of foresight.}
  \label{fig:foresight}
  \vspace{-6.5mm}
\end{figure*}
AR visual generation resembles assembling a jigsaw puzzle without seeing the full picture: each piece may fit locally, while the global structure emerges only much later. They serialize images into a sequence of discrete tokens in raster order and learn with strictly causal, one-step teacher forcing~\cite{chen2020generative, parmar2018image, child2019generating, sun2024autoregressive, NEURIPS2021_a4d92e2c, ding2022cogview2, yu2022scaling, ramesh2021zero, yu2023scaling}. While this paradigm thrives in language modeling, and limited lookahead via Multi-Token Prediction (MTP)~\cite{gloeckle2404better,liu2024deepseek} introduces further benefits, it remains ill-suited to vision data, where tokens depend on bidirectional and long-range context. As a result, global cues propagate only through many AR steps, often producing images that are locally consistent but globally misaligned. Concretely, as shown in the first image of \cref{fig:overview}, the parrot generated by the AR baseline LlamaGen-B \cite{sun2024autoregressive} exhibits an unnatural pose with a disconnected head.

We hypothesize that a missing ingredient in visual AR is the training-time foresight, \ie, signals derived from future tokens. If AR’s image representations were guided not only by the causal prefix and the immediate next token, but also by the foresight in training, the model could learn to plan ahead, forming internal states that anticipate upcoming structure while preserving causal decoding at inference. 
To validate this argument, we conduct a series of diagnostic experiments along three axes (also illustrated in \cref{fig:foresight}): 
(1)~\emph{injection level}, injecting foresight at the output \emph{vs.}\ at the internal representation level; (2)~\emph{foresight positioning}, foresight should be positioned in 1D row scan \emph{vs.} in 2D grid; (3)~\emph{source of foresight}, from implicit alignment to a bidirectional encoder or explicit alignment to a unidirectional encoder. 
Across the three axes, we discover a common pattern: injecting foresight into visual AR by aligning with its internal representations in a 2D grid yields stronger causal dependencies and a more coherent spatial organization. 
This reveals a fundamental limitation of strictly causal training: the absence of global planning signals.

This motivates our \modelname{}, a general training framework that injects future information into AR models alongside the next token prediction objective, leaving the architecture and the inference process unchanged. \modelname{} aligns the AR model's internal representations with the foresight encoded from the foresight encoder in a 2D grid. Depending on the configuration of the foresight encoder, \modelname{} admits two instantiations: \modelunidir{} provides \emph{explicit}, position–indexed foresight from the unidirectional AR model's own Exponential Moving Average (EMA), aligning internal state to the foresights at a small set of nearby future locations. In contrast, \modelbidir{} supplies \emph{implicit}, context–aggregating foresight by aligning internal states to features from a frozen bidirectional encoder at matched spatial locations. At test time, the additional alignment components are removed; decoding remains token-by-token, strictly causal, and identical in computational cost to the standard AR model.
In summary, the contributions of our paper are threefold:
\begin{itemize}
        \item We systematically investigate the effectiveness of incorporating foresight into the visual AR model and show the superiority of projecting foresight into the internal representation level over the prediction level.
        \item We propose \modelname{}, a simple yet effective framework for aligning visual AR models with 2D latent foresight. Specifically, we propose two variants of \modelname{} that utilize foresight derived from two kinds of foresight encoders.
        \item \modelname{} significantly accelerates the training of AR models and improves the quality of generated results. 
        \modelname{} can speed up LlamaGen-B's \cite{sun2024autoregressive} convergence by up to 10$\times$, and reduce the final FID from 5.34 to 4.34. 
\end{itemize}

\section{The Blessing of Foresight}

\subsection{Preliminaries}
We briefly review the formulation of AR visual generation under the discrete tokenization paradigm.
Let an image $\mathbf{X} \in \mathbb{R}^{H\times W\times 3}$ be represented by a sequence of discrete tokens $\bm{x} = [x_1, x_2, \dots, x_N]$,
where each $x_n \in \{1, \dots, V\}$ indexes a code from a learned visual vocabulary of size $V$, typically obtained from a pretrained tokenizer (\eg, VQVAE \cite{van2017neural} or VQGAN \cite{esser2021taming}).
AR models define the joint distribution over tokens as a product of conditionals:
\vspace{-4mm}
\begin{equation}
p_{\theta}(\bm{x}) = \prod_{n=1}^{N} p_{\theta}(x_n \mid \x_{<n}),
\vspace{-4mm}
\label{eq:ar_factorization}
\end{equation}
where $\x_{<n} = [x_1, \dots, x_{n-1}]$ denotes all preceding tokens.
During training, the parameters $\theta$ define an AR model $D_\theta$ and are optimized by maximizing the log-likelihood:
\vspace{-3.5mm}
\begin{equation}
\mathcal{L}_{\text{NTP}}(\theta)
= -\,\mathbb{E}_{\bm{x}\sim p_{\text{data}}}
\left[
\frac{1}{N}\sum_{n=1}^{N}
\log p_{\theta}(x_n \mid \x_{<n})
\right],
\vspace{-3mm}
\label{eq:ntp_loss}
\end{equation}
\vspace{-0.5mm}
commonly referred to as the next-token prediction (NTP) loss.
Nevertheless, the purely causal supervision in \cref{eq:ntp_loss}
provides each step with only local feedback,
which can hinder convergence speed and global coherence. This is a key motivation for introducing foresight training signals discussed in the next section.

\subsection{Autoregressive Modeling with Foresight}
\paragraph{Foresight.}
Intuitively, foresight is any auxiliary supervision that exposes the model to information about how the image will unfold beyond the immediate next token.
Consider an AR model $D_\theta$ with hidden states 
$\bm{h}_n = D_{\theta}^{[:l]}(\x_{<n})$ at position $n$ and layer $l$
(we omit $l$ when clear from context).
We call an auxiliary training signal \emph{foresight} at position $n$ if it depends on the future-side tokens $\bm{x}_{\ge n} = [x_n,\dots,x_N]$, in addition to possibly depending on the past $\bm{x}_{<n}$. We write these targets as
\vspace{-2.mm}
\begin{equation}
\label{eq:foresight}
\bm{f}_n = \{\bm{f}_n^{[k]}\}_{k = 1}^{K} = \{R(\x)_{j}:\forall j \in \mathcal{N}_K(n)\},
\vspace{-3mm}
\end{equation}
where the foresight targets $\bm{f}_n$ can be future tokens themselves or 
future-aware features, $K$ denotes the number of foresight targets per 
position $n$, and $\mathcal{N}_K(n)$ is a small set of the future position, and $R(\cdot)$ is a Foresight Encoder, parametric or not. Intuitively, a foresight objective encourages each hidden state to anticipate 
how the rest of the sequence will unfold and can be formally written as
\vspace{-2mm}
\begin{equation}
\label{eq:loss_foresight}
\mathcal{L}_{\text{Foresight}} 
= \mathbb{E} \left[ \frac{1}{NK}\sum_{n = 1}^{N} \sum_{k = 1}^{K} 
\ell\big(\bm{f}_n^{[k]}, \rho_k(\bm{h}_n)\big) \right].
\vspace{-3mm}
\end{equation}
Here, $\ell$ is a task-specific prediction loss, and $\rho_k$ is a projection 
head that maps $\bm{h}_n$ to the same dimension as $\bm{f}_n^{[k]}$.
Built upon the formulation above and the visual AR method 
LlamaGen~\cite{sun2024autoregressive}, we systematically explore the design 
space of foresight in the remainder of this subsection.

\begin{table}[t] 
\vspace{-0.5em}
\small
\centering
\caption{
\textbf{Where to inject the foresight and how the foresight is positioned}. Inj. Lvl. is short for the injection level, \ie, where foresight is applied; Layout specifies how foresight positions are chosen; $K$ is the number of foresight tokens. All experiments are performed on ImageNet 256×256 with an 80-epoch training.}
 \vspace{-3mm}
\label{tab:MTPMTA}
\begin{tabular}{l | c  c  c | c c }
\toprule
Model & Inj. Lvl. & Layout & $K$ &  FID$\downarrow$  & IS$\uparrow$ \\  
\midrule

LlamaGen-B & -- & --&  -- & 6.36  & 185.54 \\
\midrule
\multirow{9}{*}{+ Foresight}
& \multirow{2}{*}{Output} & 1D &  3 & 7.28 & 163.31 \\
&  & 2D&  3 &  6.48 & 185.57 \\

\cmidrule(lr){2-6}


& \multirow{2}{*}{Internal} & 1D &  3 & 6.20  & 176.36 \\
& & \cellcolor{color_ours}2D&  \cellcolor{color_ours}3 &  \cellcolor{color_ours}\textbf{5.22} & \cellcolor{color_ours}\textbf{197.14} \\
\cmidrule(lr){2-6}

& \multirow{2}{*}{Internal} & 1D& 4 & 6.61 & 167.87 \\
& & 2D & 4 & 5.64 & 189.20    \\
\cmidrule(lr){2-6}

& \multirow{2}{*}{Internal} &1D& 9 & 7.19  & 158.29 \\
& & 2D & 9 &  6.42 & 171.50  \\

\bottomrule

\end{tabular}
 \vspace{-2.em}
\end{table}







\paragraph{Foresight Injection Level.}
We begin by clarifying the foresight injection level, that is, where the foresight information flows into AR training. A simple instantiation is to select the next $K$ tokens as foresight, \ie, $R(\x) = \x$ and
\vspace{-3mm}
\begin{equation}
\label{eq:foresight_mtp}
\bm{f}_{n} = \{x_{n+k-1}\}_{k = 1}^{K}; \bm{h}_n = D_\theta^{[:L]}(\x_{<n}),
\vspace{-3mm}
\end{equation}
applying the generic foresight loss in \cref{eq:loss_foresight} atop the final layer $l = L$ of $D_\theta$. 
In this case, each projection head $\rho_k$ outputs a token distribution and $\ell$ becomes the cross-entropy loss as in \cref{eq:ntp_loss}. 
This design recovers Multi-Token Prediction (MTP)~\cite{gloeckle2404better,liu2024deepseek} in language modeling, which induces competing gradient cross targets and hampers optimization due to the increased difficulty. 
We conjecture that output-level foresight introduces target competition because a single hidden state must simultaneously support the immediate next-token prediction and several harder future-token predictions in the discrete token space. 

Rather than burdening the output head with multiple future-token predictions, we instead use foresight solely to supervise the model’s \emph{internal representation}, asking the model not to emit foresight tokens but to align its hidden states to them. As in \cref{fig:foresight}a, we align the model's internal representation at the $l$th layer ($0 < l < L$) to the foresight:
\vspace{-3mm}
\begin{equation}
\label{eq:foresight_mta}
\bm{f}_n = \{R_{\phi}(\x)_{n + k - 1}\}_{k = 1}^{K}, \bm{h}_n = D_\theta^{[:l]}(\x_{<n}),
\vspace{-3mm}
\end{equation}
where $D_\theta^{[:l]}$ denotes the first $l$ layers of $D_\theta$. We instantiate $R_{\phi}$ as the Exponential-Moving-Average (EMA) of $D_\theta^{[:l]}$ for simplicity and direct comparison with output prediction, updating it at each step by $\phi \leftarrow \tau \phi + (1 - \tau)\theta$ with EMA coefficient $\tau$. $\ell$ becomes the negative cosine similarity. With a foresight window of $K = 3$, we train LlamaGen-B~\cite{sun2024autoregressive} for 80 epochs and report the results in \cref{tab:MTPMTA}. Directly predicting foresight at the output level underperforms the baseline, indicating that supervising multiple discrete future tokens in a single step introduces harmful gradient interference in visual generation. By contrast, aligning at the internal level yields clear gains: aligning intermediate representations $\bm{h}_n$ to foresight $\bm{f}_n$ regularizes hidden states without predicting discrete tokens, exposes structured future information, and encourages the model $D_\theta$ to focus on next-token prediction.

\begin{figure}[t]
  \centering
    \includegraphics[width=0.65\linewidth]{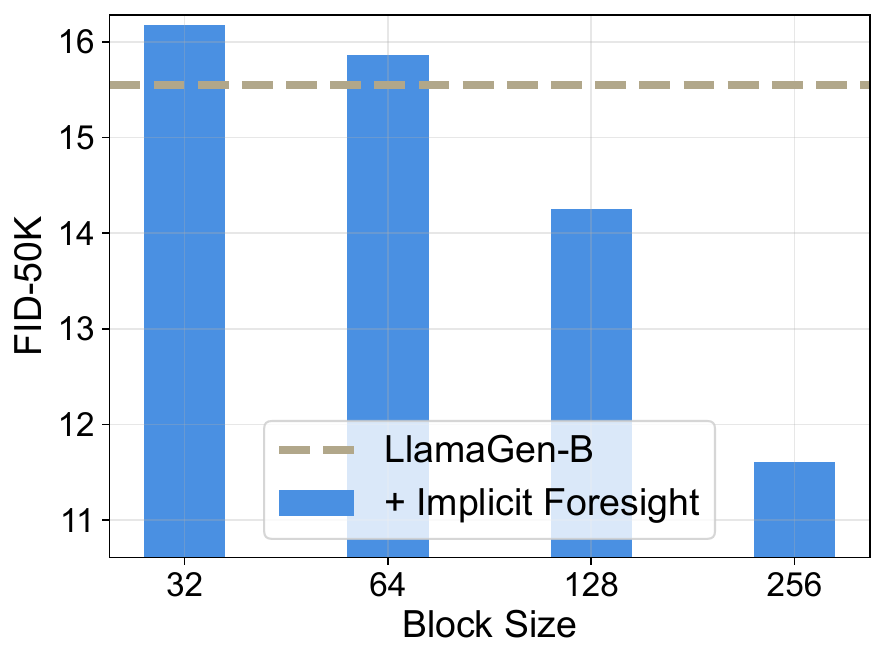}
    \vspace{-3mm}
    \caption{\textbf{Internal representation alignment with implicit foresight from bidirectional encoder}. All experiments are performed on ImageNet 256×256 with a 50k-step training.}
    \label{fig:block}
     \vspace{-7mm}
\end{figure}

\paragraph{Foresight in 1D Scan \emph{vs.}\ 2D Grid.}
We next study how foresight should be positioned in the spatial layout for visual tasks. In previous experiments, we used one-dimensional (1D) foresight, where future tokens are selected purely by raster-scan order. Formally, for a given position $n$ and window size $K$, the 1D foresight neighborhood in \cref{eq:foresight} is
\vspace{-3.5mm}
\begin{equation}
\mathcal{N}^{\text{1D}}_K(n) = \{n, n+1, \dots, n+K-1\}.
\vspace{-3mm}
\end{equation}
We now consider a two-dimensional (2D) strategy that selects foresight based on spatial nearest neighbors on the 2D image grid (see \cref{fig:foresight}b), which better reflects visual geometry. Let $q_n$ denote the 2D grid coordinate of token $x_n$; we define the 2D foresight neighborhood in \cref{eq:foresight} as the set of $K$ nearest spatial neighbors of $q_n$,
\vspace{-3mm}
\begin{equation}
\label{eq:2d_indices}
\mathcal{N}^{\text{2D}}_K(n)
= \operatorname*{arg\,topK}
\big(-\lVert q_n - q_j \rVert_2\big).
\vspace{-3mm}
\end{equation}
Using the same training setup and varying only the spatial layout, \cref{tab:MTPMTA} shows that 2D alignment consistently outperforms 1D alignment across different foresight sizes.
This result is central: in visual autoregressive modeling, the usefulness of future-aware supervision depends not only on what future information is injected, but also on how that information is positioned on the 2D token grid.
Respecting the 2D spatial structure provides more geometrically coherent foresight, encouraging the AR model to maintain consistent local neighborhoods in its internal representations.
In contrast, 1D alignment may pair spatially less relevant regions along the scan path, weakening the supervisory signal and reducing global consistency.

\begin{figure}[t]
  \centering
    \includegraphics[width=.85\linewidth]{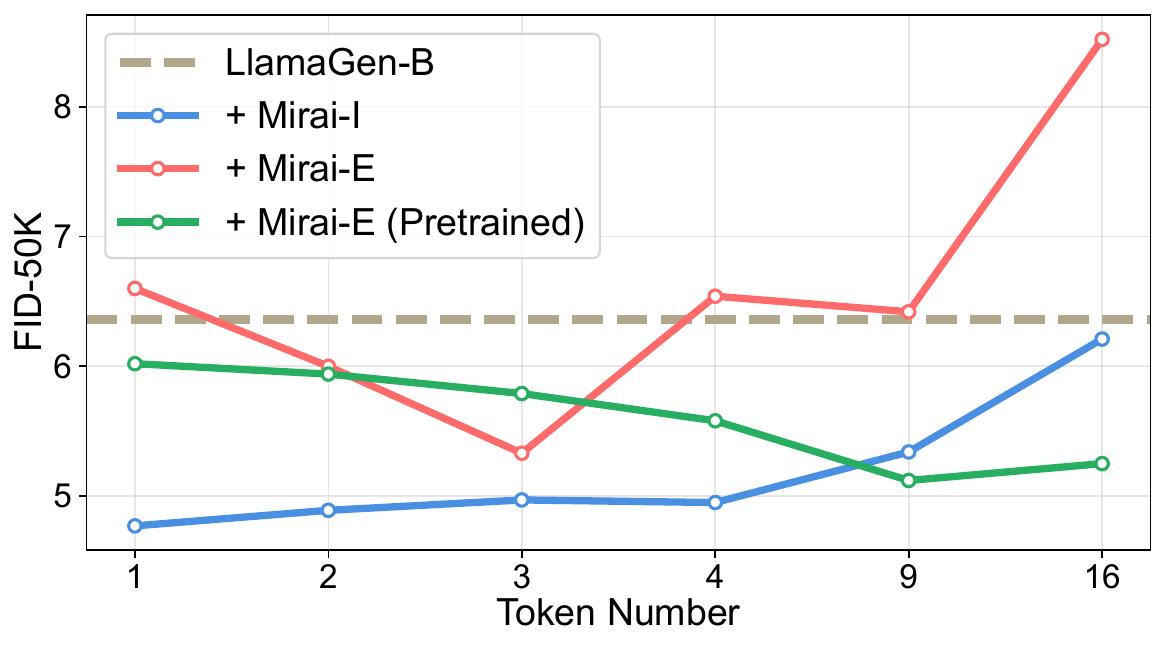}
    \vspace{-4mm}
    \caption{\textbf{Foresight token number analysis}. All models are LlamaGen-B trained for 80 epochs. } 
    \label{fig:head}
    \vspace{-2em}
\end{figure}

\paragraph{The Source of Foresight.}
So far, the foresight signal still originates from a unidirectional model and therefore preserves an explicit notion of future position. This raises a complementary question: can visual AR also benefit from a more implicit form of foresight encoded in globally contextualized bidirectional features?
We then consider an external bidirectional encoder that extracts foresight from the image $\X$, \ie,

\vspace{-6mm}
\begin{equation}
\label{eq:foresight_bidir_src}
\bm{f}_n = R_{\phi}(\X)_n,
\vspace{-2mm}
\end{equation}
where $R_{\phi}$ is now instantiated by a pretrained bidirectional vision encoder. We note that, because each of its output representations contains information about the full image, it also \emph{implicitly} embeds foresight, as shown in \cref{fig:foresight}c.

We conduct a diagnostic experiment to verify whether foresight generated by a bidirectional encoder can benefit visual AR models.
The AR model $D_{\theta}$ will align its internal states to the representation in the same position from a bidirectional encoder $R_{\phi}$, DINOv2 \cite{oquab2023dinov2}. The encoder's attention map will be restricted gradually using block-causal masking. A smaller block size limits the encoder’s ability to access future context, while a larger block restores a global view. With the same experiment setup, the results in \cref{fig:block} show a clear monotonic trend: as the encoder’s future access is reduced, generation quality degrades; restoring full bidirectional context yields the best performance and ultimately surpasses the AR baseline.
This finding reveals a key insight that the AR model can form internal representations that implicitly anticipate upcoming structure by aligning it with implicit foresight provided by a bidirectional foresight encoder. Conversely, without such foresight, the model remains locally plausible but globally fragmented.

\subsection{Methodology: \modelname{}}
\label{sec:method}

Based on the above investigations, we found that aligning foresight from either a bidirectional or unidirectional encoder to AR's intermediate representations in a 2D layout is not a violation of causality, but a catalyst for learning it. Motivated by this, we propose a family of training schemes, \modelname{}, which augment the next token prediction with foresight alignment. With the equipment of \modelname{}, the total loss function of visual AR can be written as:
\vspace{-3mm}
\begin{equation}
\mathcal{L}_{\text{\modelname{}}}
= \mathcal{L}_{\text{NTP}}
+ \lambda \mathcal{L}_{\text{Foresight}},
\vspace{-3mm}
\end{equation}
where $\mathcal{L}_{\text{NTP}}$ is the NTP loss defined in \cref{eq:ntp_loss}, $\lambda > 0$ is a hyperparameter controlling the tradeoff between next token prediction and foresight alignment, and $\mathcal{L}_{\text{Foresight}}$ denotes the foresight alignment loss, defined as: 
\vspace{-3mm}
\begin{equation}
\hspace{-3mm}
\mathcal{L}_{\text{Foresight}}
= - \mathbb{E} \left[
\frac{1}{NK} \sum_{n=1}^{N} \sum_{k=1}^{K}
\text{sim}\!\left( \bm{f}_{n}^{[k]}, \rho_{k}(\mathbf{h}_{n}) \right) \right ].
\label{eq:Mirai_loss}
\vspace{-3mm}
\end{equation}
This loss maximizes the similarities between the foresight representation $\bm{f}_n^{[k]} \in \mathbb{R}^C$ and the projection of AR model's internal representation $\rho_{k}(\bm{h}_n) \in \mathbb{R}^{C}$, where $N, C > 0$ denote the number of the output patches and the embedding dimension, respectively and $\text{sim}(\cdot,\cdot)$ denotes the cosine similarity. 
Each $\rho_{k}$ is a lightweight projection head, \eg, a multilayer perceptron (MLP), that maps the internal representation $\bm{h}_n$ into the same embedding dimension $C$ of foresight and decouples the alignment parameters from the AR backbone. During inference, the projection heads are discarded. Decoding proceeds token-by-token, remaining strictly causal and computationally identical to the baseline AR model. Depending on the source of foresight, \modelname{} has two instantiations, detailed below.

\begin{figure}[t]
  \centering
    \includegraphics[width=.75\linewidth]{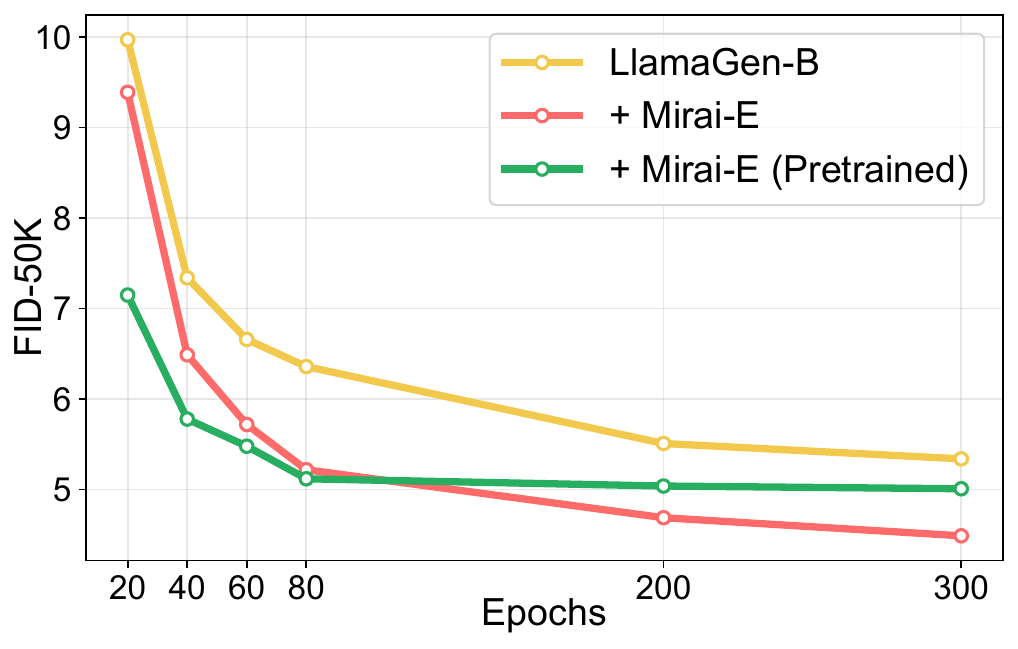}
    \vspace{-4mm}
    \caption{\textbf{Two EMA selection strategies}. All models are LlamaGen-B trained for 300 epochs. }
    \vspace{-2em}
    \label{fig:EMA}
\end{figure}

\paragraph{\modelunidir{}: Explicit Foresight.}
In \modelunidir{}, the foresight encoder $R_{\phi}$ is the EMA of the AR decoder $D_{\theta}^{[:l]}$, applied to the discrete token sequence $\bm{x}$ to produce $\bm{f}_n = \{R_{\phi}(\bm{x})_{j} : \forall j\in \mathcal{N}^{\text{2D}}_K(n)\}$, where $\mathcal{N}^{\text{2D}}_K(n)$ as in \cref{eq:2d_indices}. Because $R_{\phi}$ has a unidirectional architecture, each foresight token in $\bm{f}_n$ provides \emph{explicit} positional lookahead that is compatible with causal decoding. To capture position–specific cues, we associate an independent projection head $\rho_k$ with each neighbor index $j\in\{1,\dots,K\}$ (ordered by the distance rule on the grid). Each $\rho_k$ maps the current hidden state $\bm{h}_n$ to the representation space of the $j$th future target. We then jointly align all targets in the neighborhood. The use of distinct heads $\{\rho_k\}$ makes the supervision \emph{explicit} in space--each hidden state is matched to $K$ concretely indexed future positions rather than a single pooled or implicit signal.

\paragraph{\modelbidir{}: Implicit Foresight.} For \modelbidir{}, the foresight encoder $R_\phi$ in \cref{eq:foresight_bidir_src} is instantiated by a pretrained bidirectional encoder, which applies on the full image $\X$, yielding $\bm{f}_n = R_{\phi}(\X)_n$. Because bidirectional self-attention aggregates full-image context, each token $\bm{f}_n$ carries implicit cues about global layout and long-range dependencies. We align the AR decoder’s hidden state $\bm{h}_n$ to the co-located foresight feature $\bm{f}_n$ transformed by a lightweight projection head $\rho$ and the similarity loss in \cref{eq:Mirai_loss}, while keeping $R_{\phi}$ frozen.
This injects 2D-anchored, globally informed supervision into intermediate representations without predicting discrete future tokens, improving global coherence.

\paragraph{Methodological Differences to Prior Work.}
\modelname{} is related in implementation to representation alignment and MTP, but differs in several fundamental aspects.
(1) \modelname{} is a general framework for injecting \emph{foresight}--training-time information from future tokens--into AR modeling. By contrast, prior work such as REPA~\cite{yu2024representation} distills pretrained \emph{semantic} features of the \emph{current} image at matched positions and is designed for diffusion or other bidirectional generators. Our supervision is inherently \emph{causal in both time and position}.
(2) We focus on \emph{strictly AR} generators and make foresight explicitly two-dimensional and position-indexed on the token grid: \modelunidir{} uses a unidirectional EMA encoder to provide \emph{explicit} lookahead to a small set of future locations, while \modelbidir{} uses a bidirectional encoder to provide \emph{implicit} global context at the same spatial coordinates.
(3) We systematically study where and how to inject foresight (output \emph{vs.}\ internal layers, 1D scan \emph{vs.}\ 2D grid, implicit \emph{vs.}\ explicit) and show that straightforward designs such as output-level MTP~\cite{gloeckle2404better,liu2024deepseek} can \emph{harm} visual AR training; in contrast, \modelname{} uses either an external encoder or the model’s own EMA as a training-only foresight source, leaving the modeling and inference methods unchanged.

\begin{table}[t]
\vspace{-1em}
\small
\centering
\caption{\textbf{Which internal layer should align with the foresight}. All models are LlamaGen-B trained for 80 epochs.}
  \vspace{-3mm}
\label{tab:depth}
\begin{tabular}{l | c | c c}
\toprule
Model & Align Layer $l$ & FID$\downarrow$ & IS$\uparrow$ \\
\midrule
LlamaGen-B  & -- & 6.36 & 185.54 \\
\midrule
\multirow{4}{*}{+ \modelbidir{}}
  & 4  & 4.98 & 204.25 \\
  & 6  & 4.81 & \textbf{208.59} \\
  &\cellcolor{color_ours}8  &  \cellcolor{color_ours}\textbf{4.77} & \cellcolor{color_ours}207.34 \\
  & 10 & 5.06 & 199.01 \\
\midrule               
\multirow{5}{*}{+ \modelunidir{}}
  & 4       & 5.99 & 181.32 \\
  & 6       & 5.62 & 190.95 \\
  &  \cellcolor{color_ours}8 & \cellcolor{color_ours}\textbf{5.22} & \cellcolor{color_ours}197.14 \\
  & 10      & 5.53 & \textbf{200.21} \\
  & 8 $\to$ 6 & 6.30 & 180.54 \\
\bottomrule
\end{tabular}
     \vspace{-2em}
\end{table}

\section{Experimental Results}
\subsection{Setup}

\paragraph{Implementation details.}
Unless otherwise specified, we strictly follow the setup in LlamaGen \cite{sun2024autoregressive}. All experiments are conducted on ImageNet \cite{deng2009imagenet}, where we apply a ten-crop augmentation at 256×256 resolution, including five spatial crops (four corners and center) along with their horizontal flips, yielding ten views per image. 
In each epoch, we randomly select one view and extract its discrete codes using a pretrained VQ-GAN \cite{esser2021taming}.
We adopt the AdamW optimizer \cite{loshchilov2017decoupled} with a constant learning rate of $10^{-4}$, using a batch size of 256, and enable cosine decay only for LlamaGen-XL experiments.
For model configurations, we adopt the B, L, and XL architectures introduced in the LlamaGen paper.
During training, the EMA parameters are updated with a slow momentum $\tau = 0.9999$.
We reuse this EMA as the foresight encoder $R$ in \modelunidir{} to provide foresight after a 15-epoch warm-up.
\modelbidir{} uses DINOv2-B to provide foresight for LlamaGen-B, DINOv2-L for LlamaGen-L/-XL.
Additional details are provided in the supplementary material.

\paragraph{Evaluation.}
We evaluate generative quality using standard metrics, including Fréchet inception distance (FID) \cite{heusel2017gans}, sFID \cite{nash2021generating}, Inception Score (IS) \cite{salimans2016improved},  as well as precision and recall \cite{kynkaanniemi2019improved}. 
To ensure fair comparison with prior work, we use the official ADM TensorFlow evaluation suite \cite{dhariwal2021diffusion} with 50,000 samples and identical reference statistics.

\paragraph{Sampling.} Following LlamaGen, we discard the projection heads and employ an autoregressive sampling strategy in our \modelname{} method. We use classifier-free guidance (CFG) \cite{ho2022classifier} with a guidance scale of 2.0 for LlamaGen-B, 1.75 for LlamaGen-L/-XL. Sampling is performed at a temperature of 1.0, with top-k $=$ 0 and top-p $=$ 1.




\begin{table}[t] 
\vspace{-1em}
\small
\centering
\caption{\textbf{Alignment coefficient $\lambda$ selection}. All models are LlamaGen-B trained for 300 epochs. }
  \vspace{-3mm}
\label{tab: Dynamic_coeff}
\begin{tabular}{l |c | c |c }
\toprule
Model & Schedule &  $\lambda$ (start $\to$ end)  &  FID$\downarrow$  \\  
\midrule

LlamaGen-B &-- & --&   5.34  \\
\midrule
\multirow{9}{*}{+ \modelunidir{}}
  & \multirow{3}{*}{Const}  & $1\to1$       & 5.00 \\
  &                         & $2\to2$       & 4.96 \\
  &                         & $3\to3$       & 5.19 \\
\cmidrule(lr){2-4}
  &       \multirow{3}{*}{Step}                  & $2\to0.5$     & 4.80 \\
  &   \cellcolor{color_ours}Step     & \cellcolor{color_ours}$2\to1$       & \cellcolor{color_ours}\textbf{4.49} \\
  &                   & $2\to1\to0.5$ & 4.64 \\

\cmidrule(lr){2-4}
  &    \multirow{2}{*}{Cosine}  & $2\to0$       & 4.97 \\
  &  & $2\to1$       & 4.98 \\

\bottomrule
\end{tabular}
     \vspace{-2em}
\end{table}

\begin{figure*}[t]
  \centering
     \includegraphics[width=.8\linewidth]{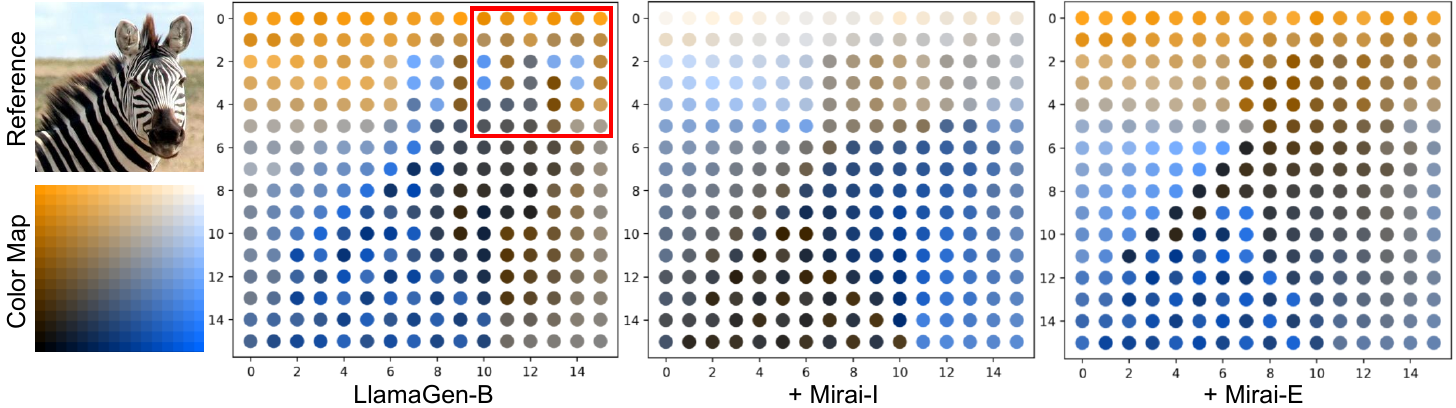}
    \vspace{-3mm}
    \caption{
    \textbf{Visualization of layer-8 internal representations on the 2D token grid}. Each token’s 2D t-SNE~\cite{maaten2008visualizing} embedding is mapped to a color (with the Color Map at bottom left) and plotted at its original grid location. Smooth color fields indicate 2D-structured representations; the red rectangle in LlamaGen-B highlights abrupt color changes where spatial structure breaks down.
    } 
    \label{fig:tsne}
    \vspace{-1em}
\end{figure*}

\begin{figure*}[t]
  \vspace{-0.1em}
  \centering
  \begin{subfigure}{0.28\linewidth}
    \includegraphics[width=\linewidth]{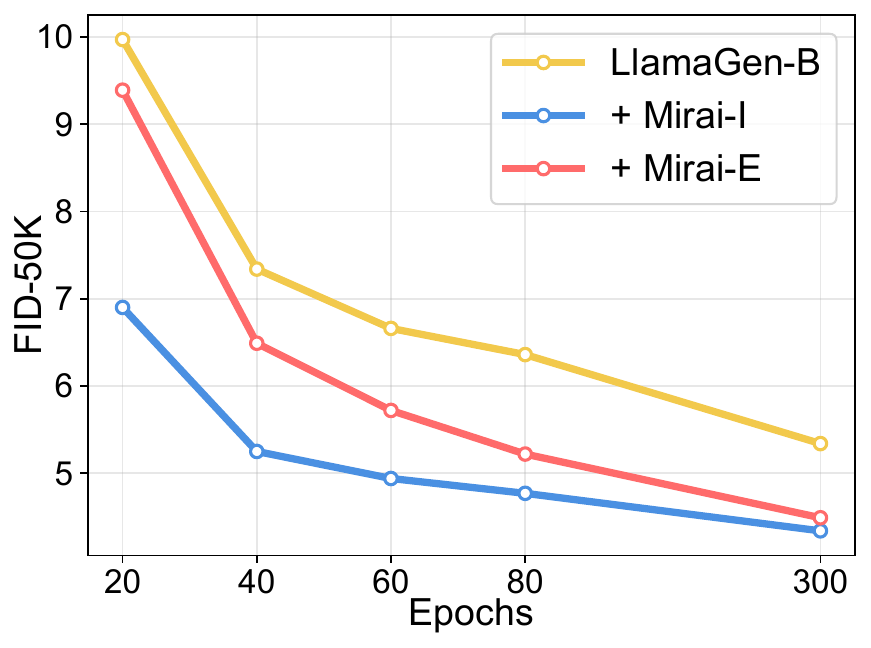}
    \label{fig:final_result_a}
  \end{subfigure}
  \hspace{1em}
  \begin{subfigure}{0.28\linewidth}
     \includegraphics[width=\linewidth]{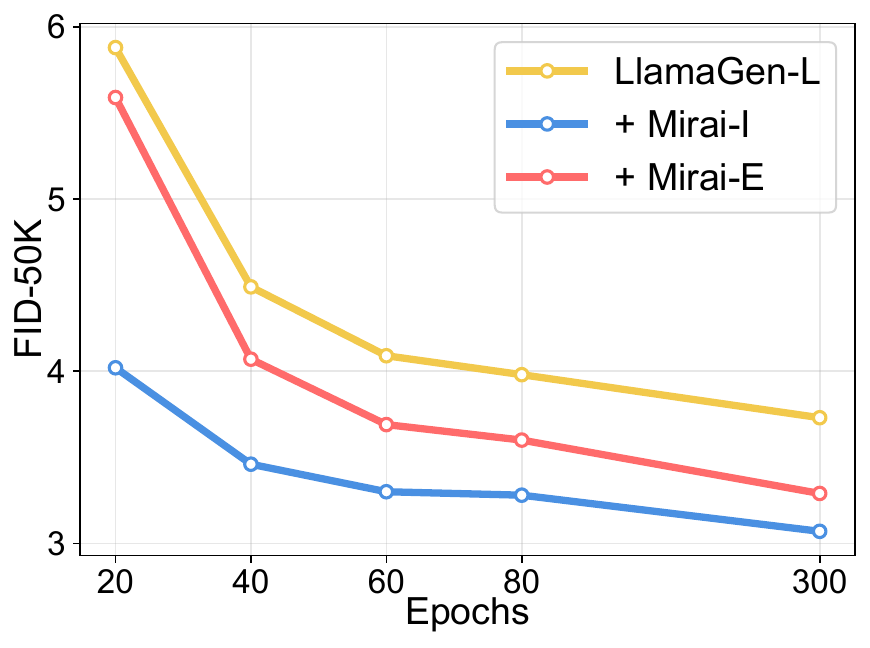}
    \label{fig:final_result_b}
  \end{subfigure}
    \hspace{1em}
  \begin{subfigure}{0.28\linewidth}
     \includegraphics[width=\linewidth]{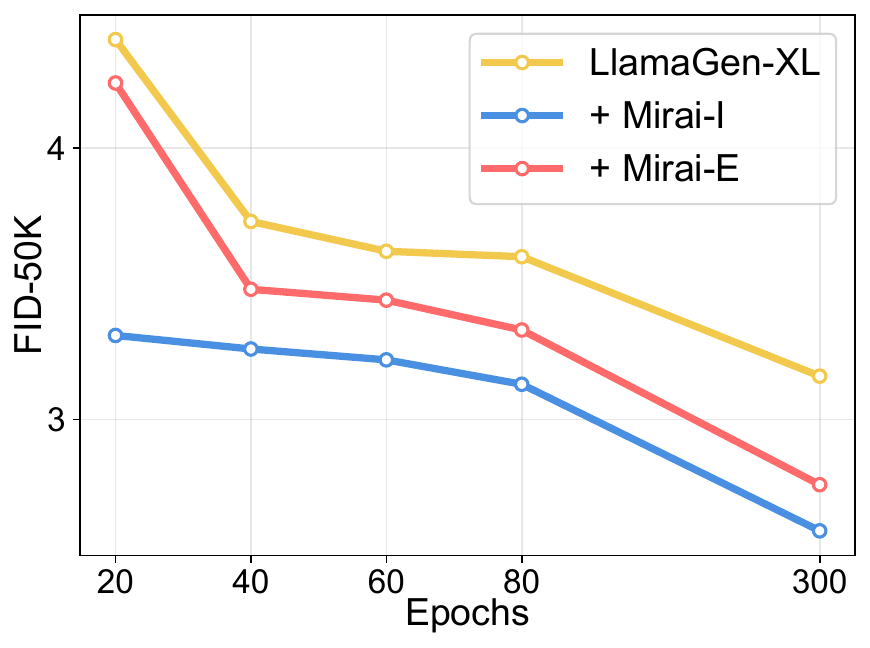}
    \label{fig:final_result_c}
  \end{subfigure}
   \vspace{-8mm}
    \caption{\textbf{FID comparisons between \modelname{}  with vanilla LlamaGen} across different model sizes and epochs on ImageNet 256×256.}
  \label{fig:final_result}
  \vspace{-6mm}
\end{figure*}

\subsection{Analysis}
\label{subsec:Component-Wise Analysis}
\paragraph{Number of Foresight Tokens.} 
We first analyze the impact of different foresight token numbers for alignment to AR's internal representation through multiple heads, with results shown in \cref{fig:head}. For \modelbidir{}, aligning only a single foresight token at the current position achieves the best performance. This stems from the bidirectional nature of DINOv2: as each of its output tokens already contains the necessary foresight, introducing extra future positions to AR would interfere with this well-learned foresight.
For \modelunidir{} with self-updated EMA, aligning 3 foresight tokens yields the best results. As the AR model and its EMA are updated jointly, aligning excessive foresight tokens may lead to conflicting gradient signals, which can hinder convergence. A moderate foresight number offers a balanced trade-off between future-aware guidance and stable optimization.

\paragraph{Instantiations of \emph{Explicit} Foresight Encoder.}
We also study \modelunidir{} with the EMA of a pretrained LlamaGen-B,  illustrated as \modelunidir{} (Pretrained). As the number of foresight tokens increases, performance peaks at 9 tokens in \cref{fig:head}. Since the pretrained EMA provides a relatively static and highly correlated supervision, more foresight tokens help the AR model capture diverse spatial offsets, leading to more comprehensive future-aware learning. 
We then compare two ways to construct EMA.  As shown in \cref{fig:EMA}, the pretrained EMA provides a stable but static supervision signal, yielding good early stage convergence but limited improvement after 80 epochs. 
In contrast, the online EMA strategy, illustrated directly as \modelunidir{}, enables stabilizing optimization with adaptive supervision.
These results indicate that an online-updated EMA provides more sustained foresight supervision than a frozen pretrained one.


\paragraph{Alignment Layer.}
We further analyze the effect of applying \modelname{} at different transformer layers of LlamaGen-B, which consists of 12 layers. As shown in \cref{tab:depth}, aligning mid-level layers, specifically the 8th layer, yields the most significant improvement in generation quality for both \modelbidir{} and \modelunidir{}. 
This indicates that intermediate layers encode semantically rich and generalizable features. This also aligns with the intuition that the lower layers primarily encode visual primitives, whereas the upper layers specialize in predicting the next token. We also attempt to align different layers when using \modelunidir{}. However, such cross-layer alignment produced the worst results, likely due to mismatched feature scales and semantic abstraction levels. 
Consequently, we apply the same relative depth ratio (8/12) when transferring to larger models in later experiments.


\paragraph{Alignment Coefficient $\lambda$.}
Then, we investigate the impact of the alignment coefficient $\lambda$, which controls the relative strength of the foresight regularization.
We use \modelunidir{} as the representative setting, as both the AR model and its EMA evolve jointly during training, making its optimization particularly sensitive to $\lambda$.
We compare three scheduling strategies: constant schedule (Const), stepwise schedule (Step), and cosine-annealing schedule (Cosine).
As summarized in \cref{tab: Dynamic_coeff}, the best performance is obtained with the step schedule that decreases $\lambda$ from 2 to 1 at the midpoint.
This indicates that maintaining strong foresight regularization is beneficial in early training to help establish global structure, while reducing its strength later helps avoid over-regularization and allows the AR model to refine token prediction.
In subsequent experiments, we adopt this step schedule as our default configuration for \modelunidir{}. \modelbidir{}'s $\lambda$ selection is provided in the supplementary material.

\begin{figure*}[t]
  \centering
    \includegraphics[width=0.82\linewidth]{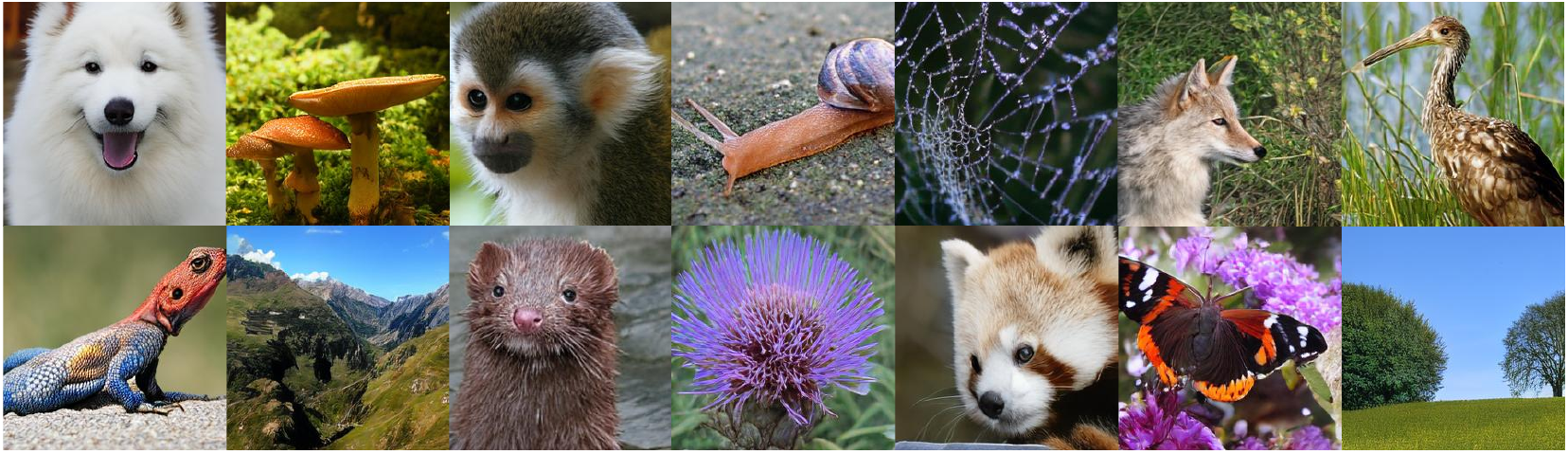}
    \vspace{-3mm}
    \caption{\textbf{Generated samples on ImageNet 256$\times$256} from the LlamaGen-XL $+$ \modelbidir{}. More results are in the supplementary material. } 
    \label{fig:sample}
    \vspace{-2em}
\end{figure*}

\paragraph{Internal Representation Visualization.}
We compute a t-SNE \cite{maaten2008visualizing} embedding of all internal representation tokens at the 8th layer for one image, then map each token’s t-SNE coordinate to a color and plot it back at its original location on the image grid.
If nearby tokens in the image share similar features, colors vary smoothly in space; if the representation ignores 2D structure, colors appear scrambled.
In \cref{fig:tsne}, compared to the LlamaGen-B, both \modelbidir{} and \modelunidir{} produce smoother, more spatially coherent color fields that align with object and background regions, indicating stronger 2D organization of internal representations. 



\paragraph{Bidirectional Foresight Encoders.}
Another study compares different bidirectional encoders providing foresight in \modelbidir{}, including DINOv2-B/-L~\cite{oquab2023dinov2}, DINOv3-B~\cite{simeoni2025dinov3}, and MAE-B/-L~\cite{he2022masked}. As shown in \cref{tab:TargetRepr}, models aligned with DINOv2-B's final representations achieve the best performance. This demonstrates that DINOv2-B's representations are more easily learned by a small size AR model. 
MAE is a pixel-reconstruction model, rather than a representation-reconstruction model, so its final outputs are not suitable for alignment. We need to obtain the foresight from its intermediate layers, specifically, the 8th layer of MAE-B (out of 12) and the 16th layer of MAE-L (out of 24). 
The alignment to the foresight from MAE leads to a marginal improvement over the baseline, suggesting that reconstruction-oriented models are less suitable for the foresight encoder. Therefore, we adopt DINOv2-B as the default foresight encoder for the LlamaGen-B with \modelbidir{}. The results on the larger size LlamaGen are provided in the supplementary material.

\begin{table} [t]
\small
\centering
\caption{\textbf{The comparison of different foresight encoders for \modelbidir{}}. All models are LlamaGen-B trained for 80 epochs.  }
 \vspace{-3mm}
\label{tab:TargetRepr}
\begin{tabular}{l | c  c | c }
\toprule
Model & Target Repr. & Enc. Params.  & FID$\downarrow$  \\  
\midrule

 LlamaGen-B  & -- &-- &6.36    \\

\midrule

\multirow{5}{*}{+ \modelbidir{}}
  & \cellcolor{color_ours}DINOv2-B & \cellcolor{color_ours}86M & \cellcolor{color_ours}\textbf{4.77}   \\

 & DINOv2-L & 300M & 4.78 \\

 & DINOv3-B & 86M &  5.02 \\

 & MAE-B &  86M & 6.34  \\
 & MAE-L  &  304M & 6.01  \\

\bottomrule
\end{tabular}
\vspace{-2.1em}
\end{table}

\paragraph{Necessity of Current Position.}
We further verify whether the current foresight token, which is at the same position as AR's internal representation token, should be aligned. The results are shown in \cref{tab: current}. 
For \modelbidir{}, aligning the current foresight token is better than aligning the next foresight token on the right.
For \modelunidir{}, we chose the best configuration in \cref{subsec:Component-Wise Analysis}, aligning the current, right, and below tokens, and compared it with the same setup but excluding the current token. \modelunidir{}'s performance when removing the current foresight token also degrades. Together, these results highlight the importance of anchoring alignment at the current spatial position to provide stable and spatially coherent foresight signals.

\subsection{System-Level Comparison}\label{sec:system-level}

We conduct a system-level comparison, comparing the FID values between vanilla Llamagen at different scales and the same models trained with \modelname{}. 
As shown in \cref{fig:final_result}, both the \modelbidir{} and \modelunidir{} consistently improve generation quality over the baselines across all scales at each training epoch. \cref{tab: System-Level} summarizes the final result of \modelname{}. Specifically, on LlamaGen-B, \modelbidir{} and \modelunidir{} achieve a reduction in FID from 5.34 to 4.34 and 4.49. The trend continues on the XL scale, where \modelbidir{} achieves a best FID of 2.59, outperforming all AR-based methods. We also compare with methods from other paradigms, including GANs, diffusion models, and masked/parallelized AR. Detailed comparisons are shown in the supplementary material.

\begin{table}[t] 
\small
\centering
\caption{\textbf{\textbf{Whether to align the current foresight token.}}  All models are LlamaGen-B trained for 80 epochs.}
 \vspace{-3mm}
\label{tab: current}
\begin{tabular}{l | c | c c}
\toprule
Model &  Aligned Token &  FID$\downarrow$  & IS$\uparrow$\\  
\midrule

LlamaGen-B & -- & 6.36& 185.54  \\
\midrule
\multirow{2}{*}{+ \modelbidir{}}
&  \cellcolor{color_ours}current &  \cellcolor{color_ours}\textbf{4.77}& \cellcolor{color_ours}\textbf{207.34} \\
 &  right & 4.99 & 202.53 \\

 \midrule

\multirow{2}{*}{+ \modelunidir{}}
 & current & 6.60 & 182.19 \\
 & right, below & 5.39 & \textbf{198.11} \\
  & \cellcolor{color_ours}current, right, below & \cellcolor{color_ours}\textbf{5.22} & \cellcolor{color_ours}197.14 \\

\bottomrule
\end{tabular}
\vspace{-2.1em}
\end{table}







\begin{table*}[t] 
\vspace{-2em}
\small
\centering
\caption{\textbf{System-Level Comparison} on ImageNet 256×256.  $\downarrow$ and $\uparrow$ indicate whether lower or higher values are better, respectively.}
 \vspace{-2mm}
\label{tab: System-Level}
\begin{tabular}{l | l  c c | c c c c c  }
\toprule
Type&  Model & Params.&   Epochs  &  FID$\downarrow$ & sFID$\downarrow$ & IS$\uparrow$ & Prec.$\uparrow$ & Rec.$\uparrow$ \\  
\midrule 
\multirow{2}{*}{GAN}
 & BigGAN \cite{brock2018large}   & 112M &--  & 6.95 & -- & 224.5 & 0.89 & 0.38 \\
& GigaGAN \cite{kang2023scaling}   & 569M & -- & 3.45 &  --  & 225.5  & 0.84 & 0.61 \\
\midrule 
\multirow{3}{*}{Diffusion}
 & LDM-4 \cite{rombach2022high}       & 400M & -- & 3.60  & -- & 247.7 &  --  &  --  \\
 & DiT-XL \cite{peebles2023scalable} & 675M &  1400  & 2.27  & 4.60 & 278.2 &  0.83  &   0.57  \\
 & SiT-XL \cite{ma2024sit} & 675M &  1400  & 2.15  & 4.50 & 258.0 &  0.81  &   0.60  \\
\midrule
\multirow{2}{*}{Mask}
 & MaskGIT~\cite{chang2022maskgit}     &227M  &  300& 6.18 & -- & 182.1 & 0.80 & 0.51 \\
 & RCG (cond.) \cite{li2023self}  & 502M & -- & 3.49 & -- & 215.5 &  --  &  --  \\

\midrule

\multirow{3}{*}{Parallelized AR}
 & VAR-d12 \cite{tian2024visual}          & 132M & -- & 5.81 & -- & 201.3  &0.81 & 0.45 \\
 & VAR-d16 \cite{tian2024visual}           & 310M& -- & 3.55 & -- & 280.4  &0.84 & 0.51 \\
 & VAR-d20 \cite{tian2024visual}           & 600M & --	 & 2.95 & -- & 302.6  &0.83 & 0.56 \\

\midrule
\multirow{3}{*}{AR}

 & VQGAN~\cite{esser2021taming}            & 1.4B & -- & 15.78 & -- & 74.3  &  --  &  --  \\
& ViT-VQGAN  \cite{yu2021vector}          & 1.7B & -- & 4.17  & -- & 175.1 &  --  &  --  \\
& RQTransformer  \cite{lee2022autoregressive}          & 3.8B &--  & 7.55  & -- & 134.0 &  --  &  --  \\
\midrule
\multirow{10}{*}{AR+\modelname{}}
&LlamaGen-B \cite{sun2024autoregressive}   & 111M & 300&  5.34 & 6.93 & 215.7 & 0.87 & 0.42 \\
&\cellcolor{color_ours}+  \modelbidir{}& \cellcolor{color_ours}111M & \cellcolor{color_ours}300& \cellcolor{color_ours}4.34 & \cellcolor{color_ours}7.13 & \cellcolor{color_ours}226.8  & \cellcolor{color_ours}0.84 & \cellcolor{color_ours}0.47\\
&\cellcolor{color_ours}+ \modelunidir{}& \cellcolor{color_ours}111M & \cellcolor{color_ours}300&\cellcolor{color_ours}4.49 & \cellcolor{color_ours}6.78 & \cellcolor{color_ours}225.7 & \cellcolor{color_ours}0.84 &\cellcolor{color_ours}0.47\\
\cmidrule(lr){2-9}

&LlamaGen-L \cite{sun2024autoregressive} & 343M&300&3.73 & 6.68 &  256.4 & 0.86 & 0.49 \\
&\cellcolor{color_ours}+ \modelbidir{} & \cellcolor{color_ours}343M& \cellcolor{color_ours}300 &\cellcolor{color_ours}3.07  & \cellcolor{color_ours}6.72  &  \cellcolor{color_ours}263.7 & \cellcolor{color_ours}0.83 & \cellcolor{color_ours}0.53 \\
&\cellcolor{color_ours}+ \modelunidir{} &\cellcolor{color_ours}343M & \cellcolor{color_ours}300&\cellcolor{color_ours}3.29 & \cellcolor{color_ours}6.64 & \cellcolor{color_ours}262.3 & \cellcolor{color_ours}0.84 & \cellcolor{color_ours}0.52\\
\cmidrule(lr){2-9}

&LlamaGen-XL \cite{sun2024autoregressive} &775M &300&3.16 & 6.55 & 293.6 & 0.85 & 0.53 \\

&\cellcolor{color_ours}+ \modelbidir{} &\cellcolor{color_ours}775M &\cellcolor{color_ours}300& \cellcolor{color_ours}2.59 & \cellcolor{color_ours}6.60 & \cellcolor{color_ours}286.9 & \cellcolor{color_ours}0.82 & \cellcolor{color_ours}0.56 \\

&\cellcolor{color_ours}+ \modelunidir{}  &\cellcolor{color_ours}775M & \cellcolor{color_ours}300&\cellcolor{color_ours}2.76 & \cellcolor{color_ours}6.48 & \cellcolor{color_ours}296.7 & \cellcolor{color_ours}0.84 & \cellcolor{color_ours}0.55 \\

\bottomrule
\end{tabular}
 \vspace{-2em}
\end{table*}

We also qualitatively compare the visual performance of generation results in \cref{fig:overview}, the model trained with \modelname{} exhibits better global consistency. As shown in \cref{fig:tsne}, \modelname{}'s nearby spatial locations form smooth color fields rather than the scrambled patterns observed in the vanilla LlamaGen-B baseline, indicating stronger 2D organization inside the transformer. \cref{fig:sample} provides more qualitative results.
\modelname{} also significantly accelerates model convergence. As shown in \cref{fig:overview}, training with only 40 epochs of \modelbidir{} or 80 epochs of \modelunidir{} already achieves FID comparable to the vanilla LlamaGen-B trained for 400 epochs. This corresponds to approximately 10$\times$ and 5$\times$ faster convergence, respectively, demonstrating that foresight alignment effectively enhances training efficiency and generation quality.

\section{Related Work}
\paragraph{Autoregressive Visual Generation.}
Early visual AR approaches~\cite{van2016pixel,van2016conditional} modeled images sequentially at the pixel level. Subsequent works like VQVAE \cite{van2017neural}, VQGAN \cite{esser2021taming}, and DALL-E \cite{ramesh2021zero} tokenize images into discrete codes. Although effective, these models still lag behind diffusion-based approaches \cite{ho2020denoising, peebles2023scalable, rombach2022high, karras2022elucidating} in image fidelity and scalability. LlamaGen~ \cite{sun2024autoregressive} advances the AR paradigm with a large-scale, pure GPT-style transformer trained over discrete image tokens. Its success demonstrates that with sufficient scale and a high-quality tokenizer, AR models can surpass diffusion models on image generation.



\vspace{-0.2em}
\paragraph{AR with Multi-token Prediction.}
Recently, there have been efforts~\cite{yang2019xlnet,du2022glm,bao2020unilmv2,ren2025beyond,yu2025randomized,yue2025understand} to go beyond AR's next token prediction paradigm by predicting multiple future tokens in a single inference. MaskGIT \cite{chang2022maskgit} follows a masked-token refinement process that predicts multiple tokens in parallel and iteratively updates low-confidence positions. Multi-Token Prediction \cite{gloeckle2404better} trains language AR to predict multiple future tokens by a shared trunk with multiple independent prediction heads. These methods yield faster sampling but sacrifice generation quality. MuToR \cite{gerontopoulos2025multi} uses register tokens to alleviate the quality degradation observed in previous multi-token prediction methods. VAR \cite{tian2024visual} redefines AR generation as next-scale prediction, generating token maps at progressively higher resolutions. While maintaining generation quality, it introduces additional pretraining costs for the encoder and decoder. In contrast, our method instead focuses on \emph{strictly autoregressive modeling} and makes foresight explicitly two-dimensional.

\vspace{-0.2em}
\paragraph{Representations for Image Generation.}
Early works explored leveraging pretrained representations to enhance the perceptual quality of generative models:
aligning latent and feature statistics between the adversarial generator and pretrained encoders was shown to stabilize training and enrich semantic consistency \cite{larsen2016autoencoding,salimans2016improved}.
Subsequent approaches capitalize on self-supervised representations as powerful semantic priors. DALL·E 2 \cite{ramesh2022hierarchical} conditions image generation on embeddings derived from a pretrained text-image encoder.
Recently, representation-aligned frameworks such as REPA \cite{yu2024representation} demonstrate that aligning intermediate generative features to pretrained encoders can substantially improve convergence and semantic coherence in diffusion transformers.
Unlike these approaches, \modelname{} explicitly studies foresight as a causality-compatible training signal in strictly autoregressive visual generation.
\vspace{-0.8em}
\section{Conclusion}
In this work, we revisited AR visual generation through the lens of foresight. We showed that purely causal supervision constrains global consistency and slows convergence. Our study revealed that foresight--signals originating from future tokens during training--can strengthen causality rather than break it. Building on this insight, we proposed \modelname{}, a general framework that injects future-aware guidance into AR training without modifying the inference architecture or increasing decoding cost. Through two instantiations, \modelbidir{} and \modelunidir{}, we demonstrated that both \emph{explicit} and \emph{implicit} foresights can accelerate convergence and enhance structural coherence. 
Comprehensive experiments on ImageNet confirm that \modelname{} substantially improves generation quality, achieving up to 10$\times$ and 5$\times$ faster convergence compared to the LlamaGen baseline. Our work highlights that autoregressive visual generation needs foresight.


\section*{Acknowledgments}
This work was partially financially supported by JST ASPIRE Program, Japan, Grant Number JPMJAP2303.
{
    \small
    \bibliographystyle{ieeenat_fullname}
    \bibliography{main}
}

\clearpage       
\appendix        

\maketitlesupplementary

\appendix

\section{More Implementation Details}
We adopt the AdamW optimizer \cite{loshchilov2017decoupled} with a constant learning rate of $10^{-4}$, using a batch size of 256, and enable cosine decay only for LlamaGen-XL experiments. $\beta$1 = 0.9, $\beta$2 = 0.95, weight decay = 0.05, gradient clipping of 1.0. The dropout is always 0.1 for the input token embedding, attention module, and FFN module. The class condition embedding dropout for classifier-free guidance is 0.1. All experiments are conducted on 8$\times$ NVIDIA A100 80GB GPUs with bfloat16 precision enabled.

\section{More Component-Wise Analysis}
\paragraph{Alignment Coefficient for \modelbidir{}.}
We investigate the impact of the alignment coefficient $\lambda$, which controls the relative strength of the foresight regularization, when using  \modelbidir{}.
We compare three scheduling strategies: constant schedule (Const), stepwise schedule (Step), and cosine-annealing schedule (Cosine).
As summarized in \cref{tab: bidir_coeff}, the constant schedule, which keeps $\lambda$ fixed at 2 for the entire training, yields the best FID.
This suggests that maintaining a relatively strong level of foresight regularization throughout is most beneficial for \modelbidir{}, stabilizing global structure through implicit foresight without preventing the AR transformer from learning fine-grained token prediction.

\begin{table} 
\small
\centering
\caption{\textbf{Alignment coefficient $\lambda$ selection for \modelbidir{}}.  All models are LlamaGen-B trained for 300 epochs. }
\vspace{-0.5em}
\label{tab: bidir_coeff}
\begin{tabular}{l |c | c |c }
\toprule
Model & Schedule &  $\lambda$ (start $\to$ end)  &  FID$\downarrow$  \\  
\midrule

LlamaGen-B &-- & --&   5.34  \\
\midrule
\multirow{5}{*}{+ \modelbidir{}}
  & \multirow{3}{*}{Const}  & $1\to1$       &  4.41 \\
  &       \cellcolor{color_ours}\multirow{1}{*}{Const}                  & \cellcolor{color_ours}$2\to2$       &  \cellcolor{color_ours}\textbf{4.34} \\
  &                         & $3\to3$       &  4.54 \\
\cmidrule(lr){2-4}
  &   \multirow{1}{*}{Step}        & $2\to1$       & 4.59 \\

\cmidrule(lr){2-4}
  &    \multirow{1}{*}{Cosine}  & $2\to0$       & 4.59 \\

\bottomrule
\end{tabular}
\vspace{-1em}
\end{table}

\paragraph{Foresight Encoder for LlamaGen-L with \modelbidir{}.} We further study the choice of foresight encoder in \modelbidir{} when scaling up the AR backbone from LlamaGen-B to LlamaGen-L.
As shown in \cref{tab: LlamaGenL_TargetRepr}, DINOv2-L outperforms DINOv2-B when used as the foresight encoder for LlamaGen-L. This indicates a scaling correspondence between the foresight encoder and the AR generator: while DINOv2-B provides the most learnable representation for the smaller size LlamaGen-B, the larger size LlamaGen-L benefits from the richer and more expressive features of DINOv2-L. 
Therefore, we adopt DINOv2-L as the default foresight encoder in \modelbidir{} for LlamaGen-L/XL.

\begin{table} [ht]
\small
\centering
\caption{\textbf{The comparison of using different foresight encoders for \modelbidir{} on the larger size model.} All models are LlamaGen-L trained for 300 epochs.  }
\vspace{-0.5em}
\label{tab: LlamaGenL_TargetRepr}
\begin{tabular}{l | c  c | c }
\toprule
Model & Target Repr. & Enc. Params.  & FID$\downarrow$  \\  
\midrule

 LlamaGen-L  & -- & -- &3.73   \\

\midrule

\multirow{2}{*}{+ \modelbidir{}}

 & DINOv2-B & 86M & 3.47 \\
   & \cellcolor{color_ours}DINOv2-L & \cellcolor{color_ours}300M & \cellcolor{color_ours}\textbf{3.07}   \\



\bottomrule
\end{tabular}
\vspace{-1em}
\end{table}

\paragraph{Different Implementations of Projection Heads.}
We further compare two designs for the projection head $\rho_{k}$.
The first is a lightweight three-layer MLP, which maps the AR hidden state to an intermediate projector space, applies a SiLU nonlinearity, repeats this Linear–SiLU block once more, and finally projects to the foresight dimension $C$. This design introduces about 7.34M parameters.
The second variant replaces the simple MLP projector with a lightweight transformer-style block. It first applies LayerNorm followed by a 4-head self-attention layer, and adds the residual connection to preserve the original token information. A second LayerNorm–MLP block with expansion ratio 4.0 further refines token representations, followed by another residual connection. Finally, a linear projection maps the hidden dimension to $C$. This design introduces about 7.68M parameters and allows the projector to re-contextualize tokens before alignment.

As summarized in \cref{tab: head}, the transformer projector consistently underperforms the simple MLP projector for both \modelbidir{} and \modelunidir{}, despite its higher capacity.
We hypothesize that the transformer head tends to solve the foresight alignment objective within the projector itself. Due to self–attention, it can reconstruct foresight by mixing information across tokens even when the AR backbone representations are suboptimal, causing the foresight loss to be absorbed by the head rather than propagated as a strong constraint on the AR states.
In contrast, the lightweight MLP performs strictly point–wise mapping, forcing each AR token representation to carry the necessary semantic signal, which leads to more effective regularization of the backbone and better generative performance.


\paragraph{Warm-up in \modelunidir{}.} \modelunidir{} relies on the model’s own EMA as the foresight encoder. However, in the early training stage, the EMA is not yet more stable than the AR model until 15 epochs.
To avoid injecting unreliable foresight, we introduce a 15-epoch warm-up stage in which \modelunidir{} is trained only with the standard AR next token prediction, and explicit foresight is activated afterward.
As shown in \cref{tab: warmup}, adding warm-up significantly improves FID compared with applying foresight from the beginning, confirming the necessity of delaying explicit foresight alignment until the EMA becomes reliable.

\begin{table} [h]
\small
\centering
\caption{\textbf{The comparison of using different types of projection heads}. All models are LlamaGen-B trained for 80 epochs.  }
\vspace{-0.5em}
\label{tab: head}
\begin{tabular}{l |  c c | c }
\toprule
Model & Type  &  Head Params.  & FID$\downarrow$  \\  
\midrule

 LlamaGen-B  &  -- & -- & 6.36  \\

\midrule

\multirow{2}{*}{+ \modelbidir{}}

 & \cellcolor{color_ours}MLP  & \cellcolor{color_ours}7.34M & \cellcolor{color_ours}\textbf{4.77} \\ 

    & transformer  & 7.68M & 5.65 \\

\midrule

\multirow{2}{*}{+ \modelunidir{}}

 & \cellcolor{color_ours}MLP  & \cellcolor{color_ours}7.34M & \cellcolor{color_ours}\textbf{5.22} \\

     & transformer  & 7.68M & 6.80 \\


\bottomrule
\end{tabular}
\vspace{-0.5em}
\end{table}

\begin{table} [ht]
\small
\centering
\caption{\textbf{The comparison of whether using warm-up for \modelunidir{}}. All models are LlamaGen-B trained for 80 epochs.  }
\vspace{-0.5em}
\label{tab: warmup}
\begin{tabular}{l |   c | c }
\toprule
Model & Warm-up    & FID$\downarrow$  \\  
\midrule

 LlamaGen-B  & -- &6.36  \\

\midrule

\multirow{2}{*}{+ \modelunidir{}}

 & no  &  8.32 \\ 

    & \cellcolor{color_ours}yes  & \cellcolor{color_ours}\textbf{5.22}   \\


\bottomrule
\end{tabular}
\vspace{-1em}
\end{table}

\begin{table} [ht]
\small
\centering
\caption{\textbf{Mirai at 384×384 Resolution.} The generated images are 384×384 and is resized to 256×256. All models are trained for 80 epochs.  $\downarrow$ and $\uparrow$ indicate whether lower or higher values are better, respectively.}
\vspace{-0.5em}
\label{tab: 384}
\begin{tabular}{l | c  c c c c  }
\toprule
Model &   FID$\downarrow$ & sFID$\downarrow$ & IS$\uparrow$ & Prec.$\uparrow$ & Rec.$\uparrow$ \\  
\midrule

LlamaGen-B &  7.43 & 6.60  & 153.41 & 0.84 & 0.40 \\
\midrule

+  \cellcolor{color_ours}\modelbidir{}& \cellcolor{color_ours}4.91  & \cellcolor{color_ours}6.41  &  \cellcolor{color_ours}192.28 & \cellcolor{color_ours}0.83 & \cellcolor{color_ours}0.47 \\
+ \cellcolor{color_ours}\modelunidir{}& \cellcolor{color_ours}5.72 & \cellcolor{color_ours}6.33 &  \cellcolor{color_ours}182.24 & \cellcolor{color_ours}0.80 & \cellcolor{color_ours}0.45 \\

\bottomrule
\end{tabular}
\vspace{-0.5em}
\end{table}

\section{\modelname{} at Different Resolutions}
To evaluate the scalability of \modelname{} beyond 256×256 resolution, we further apply foresight alignment to 384×384 image generation on ImageNet. For evaluation, the generated images are downsampled to 256×256. The results are summarized in \cref{tab: 384}. Both \modelbidir{} and \modelunidir{} consistently improve the baseline across multiple metrics. These results suggest that \modelname{} remains effective when scaling to higher resolutions.

\section{\modelname{} on Larger Scale Models }
We further scale \modelname{} to LlamaGen-XXL (1.4B). 
As shown in \cref{tab: XXL}, foresight continues to yield improvements at the billion-parameter scale.

\section{\modelname{} on Other AR Architectures}
To further examine the generality of \modelname{}, we apply our foresight alignment strategy to a different AR architecture, Parallelized Autoregressive Visual Generation (PAR) \cite{wang2025parallelized}.
Unlike LlamaGen, which follows a strictly sequential next-token decoding process, PAR parallelizes AR inference by predicting multiple tokens in each step while preserving the causal dependency structure. This presents a different modeling bias from sequential AR and therefore serves as a strong testbed for validating the universality of our method.
As summarized in \cref{tab: PAR}, both \modelbidir{} and \modelunidir{} consistently improve PAR across most major metrics. These results show that \modelname{} is not tailored to a specific AR architecture but generalizes to AR models with different paradigms, demonstrating that foresight-based alignment is broadly applicable for enhancing AR visual generation.

\begin{table} 
\small
\centering
\setlength{\tabcolsep}{5.5pt}   
\caption{\textbf{\modelname{} on Larger Scale Models.} All models are LlamaGen-XXL trained for 40 epochs.  $\downarrow$ and $\uparrow$ indicate whether lower or higher values are better, respectively.}
\vspace{-0.5em}
\label{tab: XXL}
\begin{tabular}{l | c c c c c  }
\toprule
Model &   FID$\downarrow$ & sFID$\downarrow$ & IS$\uparrow$ & Prec.$\uparrow$ & Rec.$\uparrow$ \\  
\midrule

LlamaGen-XXL  &3.49 & 6.66 & 272.98 & 0.86 & 0.49 \\
\midrule

+ \cellcolor{color_ours}\modelbidir{}
  & \cellcolor{color_ours}3.04
  & \cellcolor{color_ours}6.73
  & \cellcolor{color_ours}284.19
  & \cellcolor{color_ours}0.84
  & \cellcolor{color_ours}0.53 \\
+ \cellcolor{color_ours}\modelunidir{}
  & \cellcolor{color_ours}3.10
  & \cellcolor{color_ours}6.65
  & \cellcolor{color_ours}289.10
  & \cellcolor{color_ours}0.85
  & \cellcolor{color_ours}0.50 \\

\bottomrule
\end{tabular}
\vspace{-0.5em}
\end{table}

\begin{table} 
\small
\centering
\caption{\textbf{\modelname{} on another AR architecture.} All models are PAR-B trained for 80 epochs.  $\downarrow$ and $\uparrow$ indicate whether lower or higher values are better, respectively.}
\vspace{-0.5em}
\label{tab: PAR}
\begin{tabular}{l | c  c c c c  }
\toprule
Model &   FID$\downarrow$ & sFID$\downarrow$ & IS$\uparrow$ & Prec.$\uparrow$ & Rec.$\uparrow$ \\  
\midrule

PAR-B & 7.47 & 7.04 & 183.60 & 0.87 & 0.36 \\

\midrule
+  \cellcolor{color_ours}\modelbidir{}&  \cellcolor{color_ours}5.59 & \cellcolor{color_ours}7.01 & \cellcolor{color_ours}201.42  & \cellcolor{color_ours}0.84 & \cellcolor{color_ours}0.44 \\
+ \cellcolor{color_ours}\modelunidir{}& \cellcolor{color_ours}6.64 & \cellcolor{color_ours}6.96 & \cellcolor{color_ours}193.13 & \cellcolor{color_ours}0.85 & \cellcolor{color_ours}0.40 \\



\bottomrule
\end{tabular}
\vspace{-1em}
\end{table}

\section{\modelname{} in Low-Resource and Limited-Data Settings}
To evaluate \modelname{} in low-resource and Limited-data scenarios, we construct a smaller model based on LlamaGen-B by halving its parameters, which we denote as LlamaGen-S. 
We train LlamaGen-S with \modelname{} on a subset of ImageNet containing 100 images per class using a single A100 GPU. 
Results in \cref{tab: samll-size} show that our method still provides improvements even in low-resource and Limited-data settings.

\begin{table} 
\small
\centering
\setlength{\tabcolsep}{5.5pt}   
\caption{\textbf{\modelname{} in low-resource and data settings.} All models are LlamaGen-S trained for 80 epochs on \textbf{1/10} ImageNet.  $\downarrow$ and $\uparrow$ indicate whether lower or higher values are better, respectively.}
\vspace{-0.5em}
\label{tab: samll-size}
\begin{tabular}{l | c c c c c}
\toprule
Model &   FID$\downarrow$ & sFID$\downarrow$ & IS$\uparrow$ & Prec.$\uparrow$ & Rec.$\uparrow$ \\  
\midrule

LlamaGen-S & 47.81 & 10.39 & 24.31 & 0.43 & 0.45 \\
\midrule

+ \cellcolor{color_ours}\modelbidir{}
  & \cellcolor{color_ours}35.62
  & \cellcolor{color_ours}9.99
  & \cellcolor{color_ours}37.66
  & \cellcolor{color_ours}0.55
  & \cellcolor{color_ours}0.45 \\
+ \cellcolor{color_ours}\modelunidir{}
  & \cellcolor{color_ours}41.84
  & \cellcolor{color_ours}9.35
  & \cellcolor{color_ours}29.60
  & \cellcolor{color_ours}0.49
  & \cellcolor{color_ours}0.45 \\

\bottomrule
\end{tabular}
\vspace{-0.5em}
\end{table}

\begin{figure*}[t]
  \centering
     \includegraphics[width=1\linewidth]{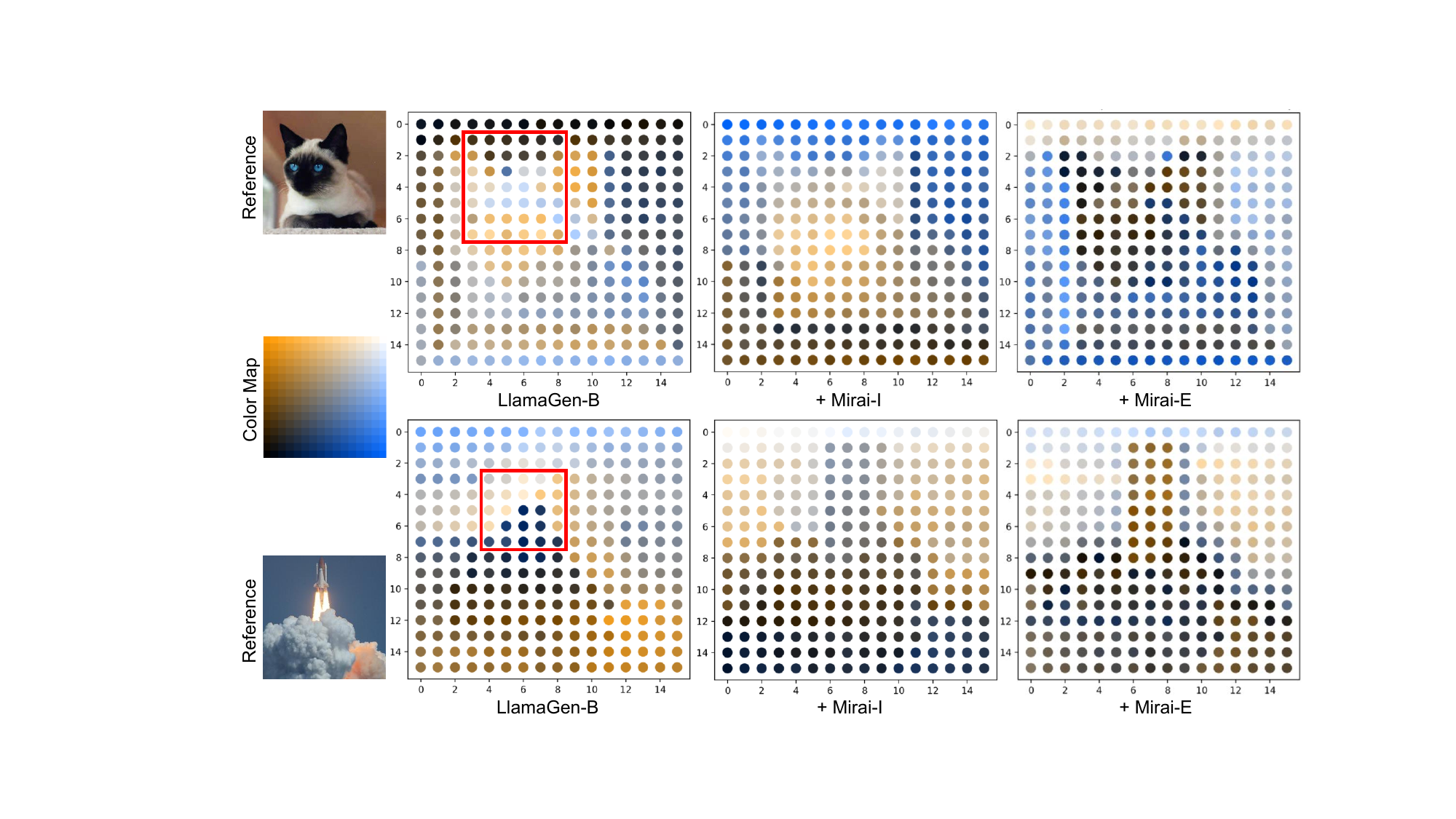}
     \vspace{-1.5em}
    \caption{
    \textbf{More results for visualization of layer-8 internal representations on the 2D token grid.} Each token’s 2D t-SNE~\cite{maaten2008visualizing} embedding is mapped to a color (with the Color Map at the middle left) and plotted at its original grid location. Smooth color fields indicate 2D-structured representations; the red rectangle in LlamaGen-B highlights abrupt color changes where spatial structure breaks down.
    } 
    
    \label{fig:tsne_app}
\end{figure*}

\section{More Internal Representation Visualization.}

We provide more internal representation visualization results by computing a t-SNE \cite{maaten2008visualizing} embedding of all internal representation tokens at the 8th layer for one image, then mapping each token’s t-SNE coordinate to a color and plotting it back at its original location on the image grid.
If nearby tokens in the image share similar features, colors vary smoothly in space; if the representation ignores 2D structure, colors appear scrambled.
In \cref{fig:tsne_app}, compared to LlamaGen-B, both \modelbidir{} and \modelunidir{} produce smoother, more spatially coherent color fields that align with object and background regions, indicating stronger 2D organization of internal representations.

\section{Evaluation Metrics}
We strictly follow the ADM suite \cite{dhariwal2021diffusion}  for evaluation and adopt their released reference batches to ensure fair comparison. We use 8× NVIDIA A100 80GB GPUs for evaluation with a batch size of 256 and enable bfloat16 for faster sampling. 
Below, we briefly summarize the evaluation metrics used in our experiments:

• \textbf{FID} \cite{heusel2017gans} quantifies the distributional discrepancy between real and generated samples by comparing Inception-V3 \cite{szegedy2016rethinking} features under the Gaussian assumption.

• \textbf{sFID} \cite{nash2021generating} extends FID by using intermediate spatial features of Inception-V3, making the metric sensitive to spatial structure in generated images.

• \textbf{IS} \cite{salimans2016improved} measures both image quality and class diversity by evaluating the KL divergence between the marginal label distribution and the conditional label distribution obtained from the Inception-V3 classifier.

• \textbf{Precision \& Recall} \cite{kynkaanniemi2019improved} separately evaluate the realism of generated samples (precision) and the coverage of the real data manifold (recall).

\section{FLOPs}
We estimate the training compute in floating point operations (FLOPs) for LlamaGen-B and \modelname{} on a per-image basis, counting only dense matrix multiplications in the transformer and projection heads while ignoring cheaper element-wise operations (e.g., LayerNorm, activations, softmax). The results are summarized in \cref{tab: FLOPs}. Relative to the LlamaGen-B baseline, \modelbidir{} increases the per-image training compute by only 6.6\%, while \modelunidir{} increases it by 38.2\%. To fairly compare training efficiency, we combine the per-image FLOP factors with convergence speed. Empirically, LlamaGen-B requires 400 epochs to reach an FID, while \modelbidir{} and \modelunidir{} converge in only 40 and 80 epochs, respectively. On an epoch basis, this corresponds to 10$\times$ and 5$\times$ faster convergence. After accounting for the FLOPs, \modelbidir{} achieves a 9.4$\times$ reduction in total training compute to reach the same FID, and \modelunidir{} achieves a 3.6$\times$ reduction.

\begin{table} 
\small
\centering
\vspace{-0.1em}
\caption{\textbf{Per-image training FLOPs and relative compute overhead.} 
FLOPs measures the per-image training cost; Compute Overhead measures the percentage increase in per-image computational cost introduced by \modelname{} relative to LlamaGen-B.}
\vspace{-0.5em}
\label{tab: FLOPs}



\begin{tabular}{l | c c      }
\toprule
Model & FLOPs  &     Compute Overhead (\%)  \\  
\midrule

LlamaGen-B & 1.70 $\times$ 10$^{11}$ &   --\\

\midrule
+  \modelbidir{}& 1.81 $\times$ 10$^{11}$   & 6.6\% \\

+ \modelunidir{}& 2.35 $\times$ 10$^{11}$   &38.2\% \\

\bottomrule
\end{tabular}
\vspace{-1em}
\end{table}








\section{Methods for Comparison}
We briefly describe the models used in the system-level comparison.

• \textbf{BigGAN} \cite{brock2018large} A class-conditional large-scale GAN that jointly scales the generator and discriminator in both capacity and resolution. Strong spectral normalization, hinge loss, and architectural improvements, \eg, shared embeddings, enable competitive FID and IS on ImageNet.

• \textbf{GigaGAN} \cite{kang2023scaling}  A high-capacity adversarial generator trained at hundreds of millions of parameters with multi-scale training and perceptual as well as CLIP-based losses. Extensive augmentation and optimization heuristics further improve GAN fidelity and IS.

• \textbf{LDM-4} \cite{rombach2022high} A latent diffusion model trained and sampled in a VAE latent space with 4× downsampling, reducing computational cost while retaining strong perceptual quality. The model employs classifier-free guidance and decodes latents back to pixel space.

• \textbf{DiT-XL} \cite{peebles2023scalable} A pure-transformer diffusion backbone that uses AdaLN-Zero conditioning and large-batch training to achieve stable scalability. The architecture demonstrates that ViT-style blocks can replace U-Nets for high-resolution diffusion.

• \textbf{SiT-XL} \cite{ma2024sit} A continuous-time, flow-style reformulation of DiT that simplifies training schedules and objectives. The model achieves higher throughput while preserving or improving image quality relative to discrete-time diffusion transformers.

• \textbf{MaskGIT} \cite{chang2022maskgit} A non-AR masked-token predictor that performs iterative parallel decoding. A bidirectional transformer fills masked codes in a few refinement steps, offering faster generation than strictly causal AR models.

• \textbf{RCG (cond.)} \cite{li2023self} Representation-Conditioned Generation that drives an image generator using self-supervised visual features instead of human labels. In our setting, RCG operates on top of a MaskGIT-style parallel decoder with additional class conditioning to refine masked-token synthesis.

• \textbf{VAR-d12/d16/d20} \cite{tian2024visual} Visual Autoregressive modeling that redefines AR learning as coarse-to-fine next-scale prediction: each higher-resolution token map is generated in parallel, conditioned on all previous scales with a block-wise causal mask. Depth tags, d12/d16/d20, denote the number of transformer layers.

• \textbf{VQGAN} \cite{esser2021taming} An AR generator that operates over discrete codes obtained from a VQGAN tokenizer and reconstructs images through its decoder. The final fidelity is constrained by the reconstruction capability of the tokenizer since the AR head only predicts token sequences.

• \textbf{ViT-VQGAN} \cite{yu2021vector} A VQGAN variant that replaces CNN modules in the tokenizer and decoder with ViT-style components, improving reconstruction fidelity and narrowing the performance gap between token reconstruction and AR generation.

• \textbf{RQTransformer} \cite{lee2022autoregressive} An AR model over residual vector-quantized tokens produced by multiple stacked codebooks. Each codebook is predicted sequentially, progressively refining quantization residuals and supporting higher-fidelity generation.

\section{Limitation}

\begin{figure}[t]
  \centering

  \begin{subfigure}{0.24\linewidth}
    \centering
    \includegraphics[width=0.9\linewidth]{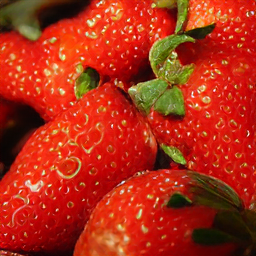}
    \label{fig:failure_cases_a}
  \end{subfigure}
  \hfill
  \begin{subfigure}{0.24\linewidth}
    \centering
    \includegraphics[width=0.9\linewidth]{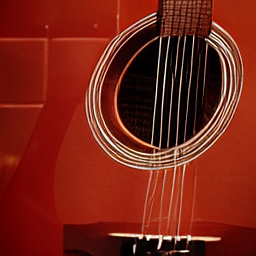}
    \label{fig:failure_cases_b}
  \end{subfigure}
  \hfill
  \begin{subfigure}{0.24\linewidth}
    \centering
    \includegraphics[width=0.9\linewidth]{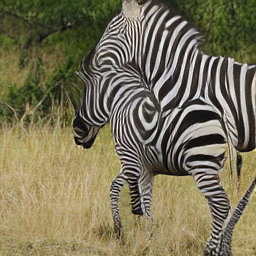}
    \label{fig:failure_cases_c}
  \end{subfigure}
    \hfill
  \begin{subfigure}{0.24\linewidth}
    \centering
    \includegraphics[width=0.9\linewidth]{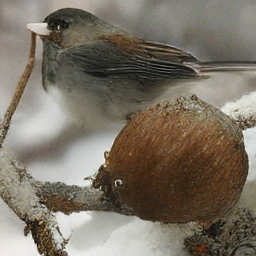}
    \label{fig:failure_cases_d}
  \end{subfigure}
\caption{\textbf{Failure cases with 16 foresight tokens.} Objects merge into the background or neighboring ones.}
  \label{fig:failure_cases}
  \vspace{-6mm}
\end{figure}

Injecting excessive foresight into \modelunidir{} can lead to failure cases, as shown in \cref{fig:failure_cases}. In particular, when the number of foresight tokens increases to 16, generated objects may partially blend into the background or merge with nearby objects. This suggests that excessive foresight may over-constrain the representation and compromise locality. Reducing the amount of foresight mitigates this issue. A promising direction for future work is to explore how to better exploit richer foresight signals without sacrificing local fidelity.

\section{Ethical Considerations}
This work improves the training of AR generation models using foresight. All experiments are conducted on the publicly available ImageNet dataset, and no private or sensitive data are used. Similar to other generative models, our approach could potentially be misused to synthesize misleading visual content. We encourage responsible use of generative technologies and acknowledge that dataset biases may be inherited from the training data.

\section{More Qualitative Results}
\begin{figure}[ht]
  \centering
     \includegraphics[width=1\linewidth]{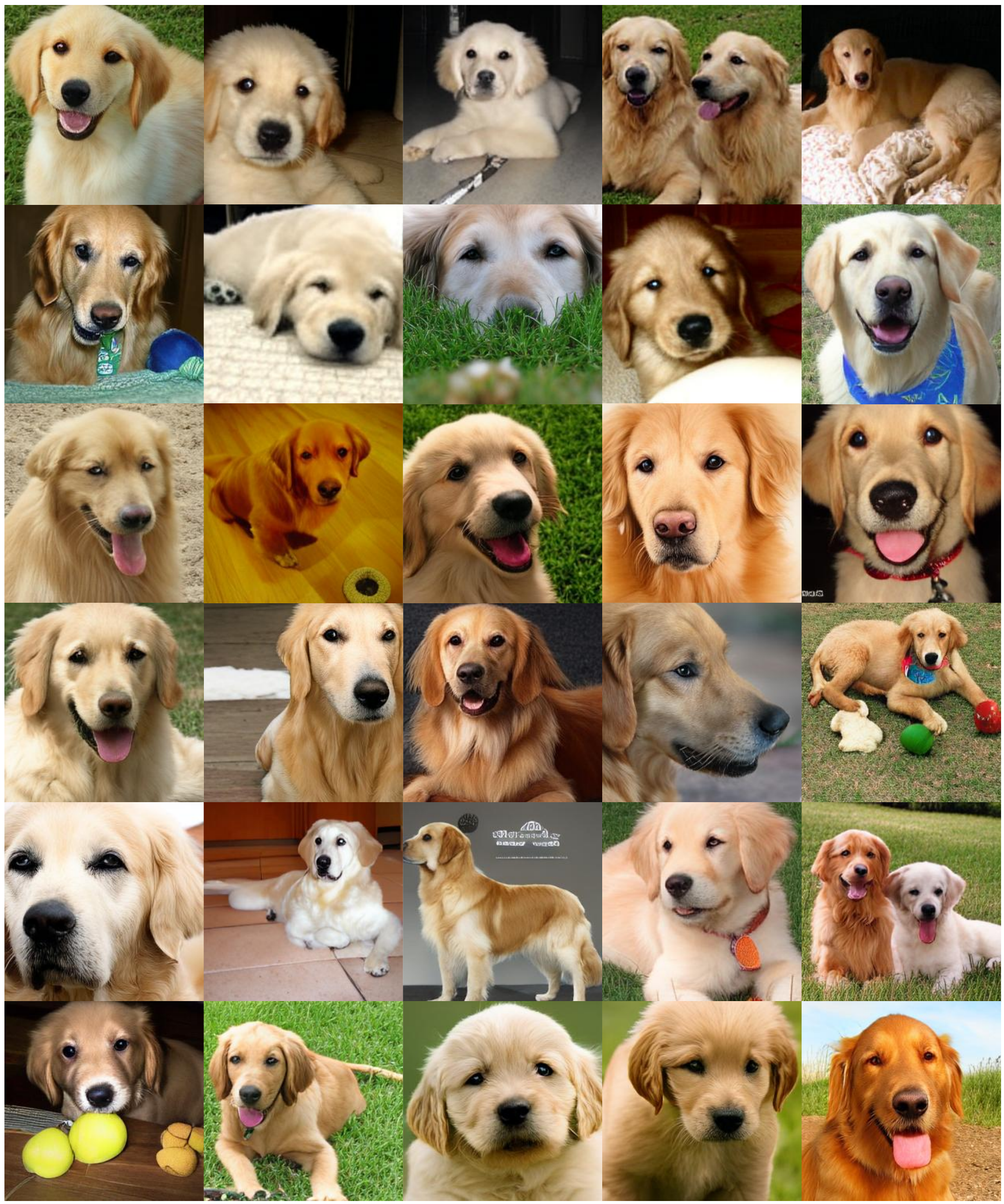}
    \caption{   256×256 LlamaGen-XL + \modelbidir{} samples. Classifier-free guidance scale = 1.75. Class label = ``golden retriever''  (207).    } 
    \label{fig:dog I}
\end{figure}

\begin{figure}[ht]
  \centering

     \includegraphics[width=1\linewidth]{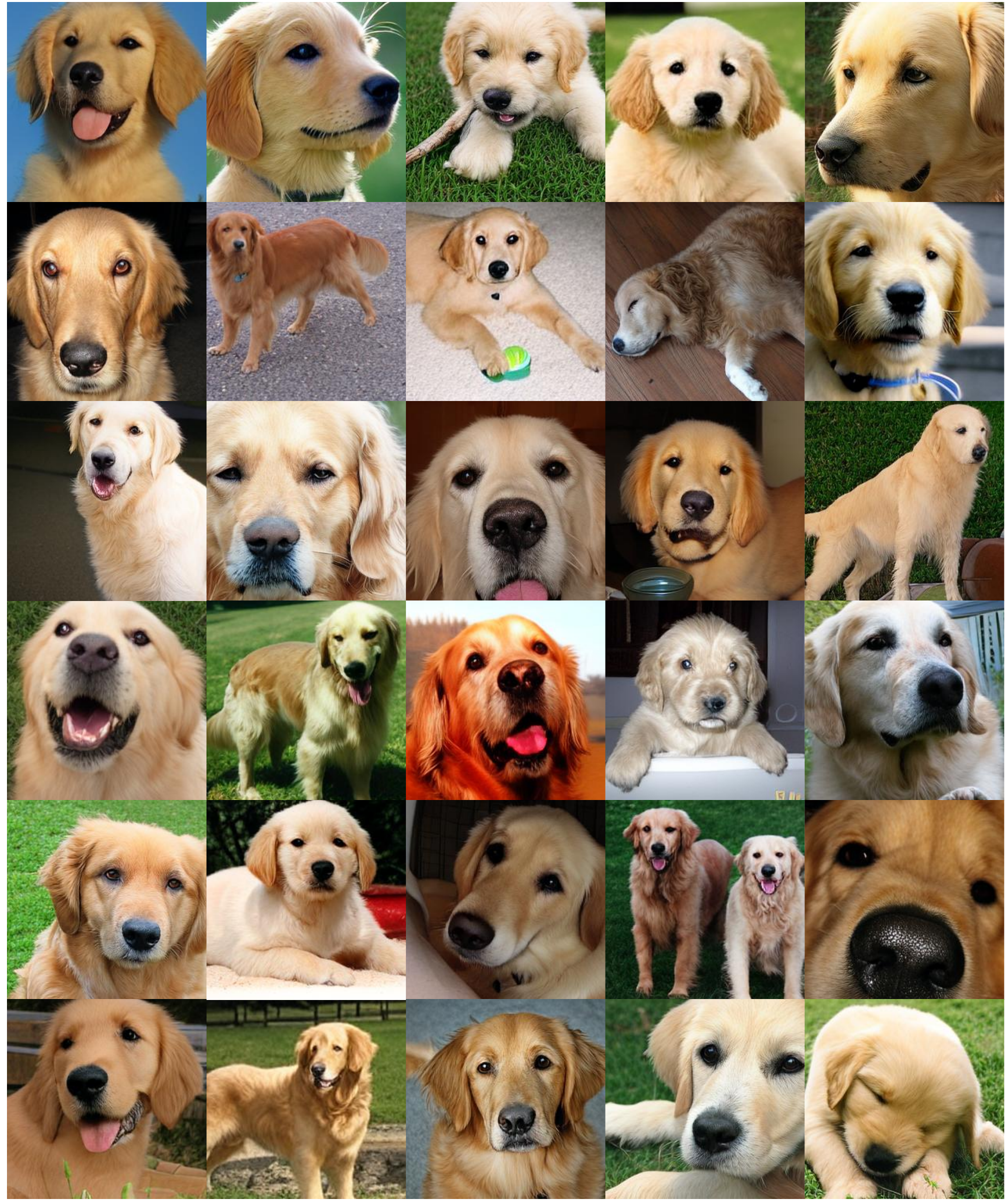}
    \caption{   256×256 LlamaGen-XL + \modelunidir{} samples. Classifier-free guidance scale = 1.75. Class label = ``golden retriever''  (207).    } 
    \label{fig:dog E}
\end{figure}

\begin{figure}[ht]
  \centering
     \includegraphics[width=1\linewidth]{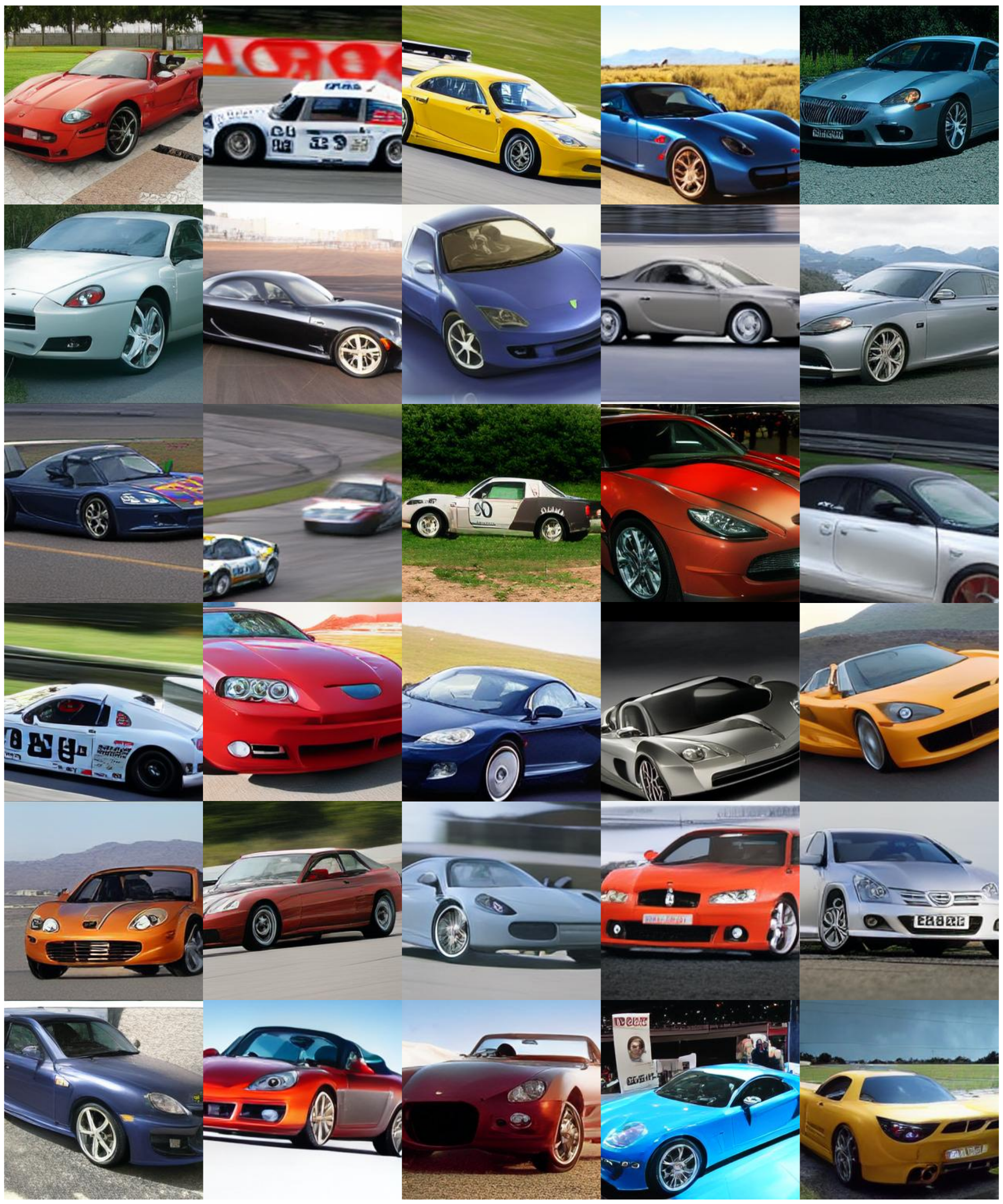}
    \caption{   256×256 LlamaGen-XL + \modelbidir{} samples. Classifier-free guidance scale = 1.75.  Class label = ``sport car'' (817).   } 
    \label{fig:car I}
\end{figure}

\begin{figure}[ht]
  \centering

     \includegraphics[width=1\linewidth]{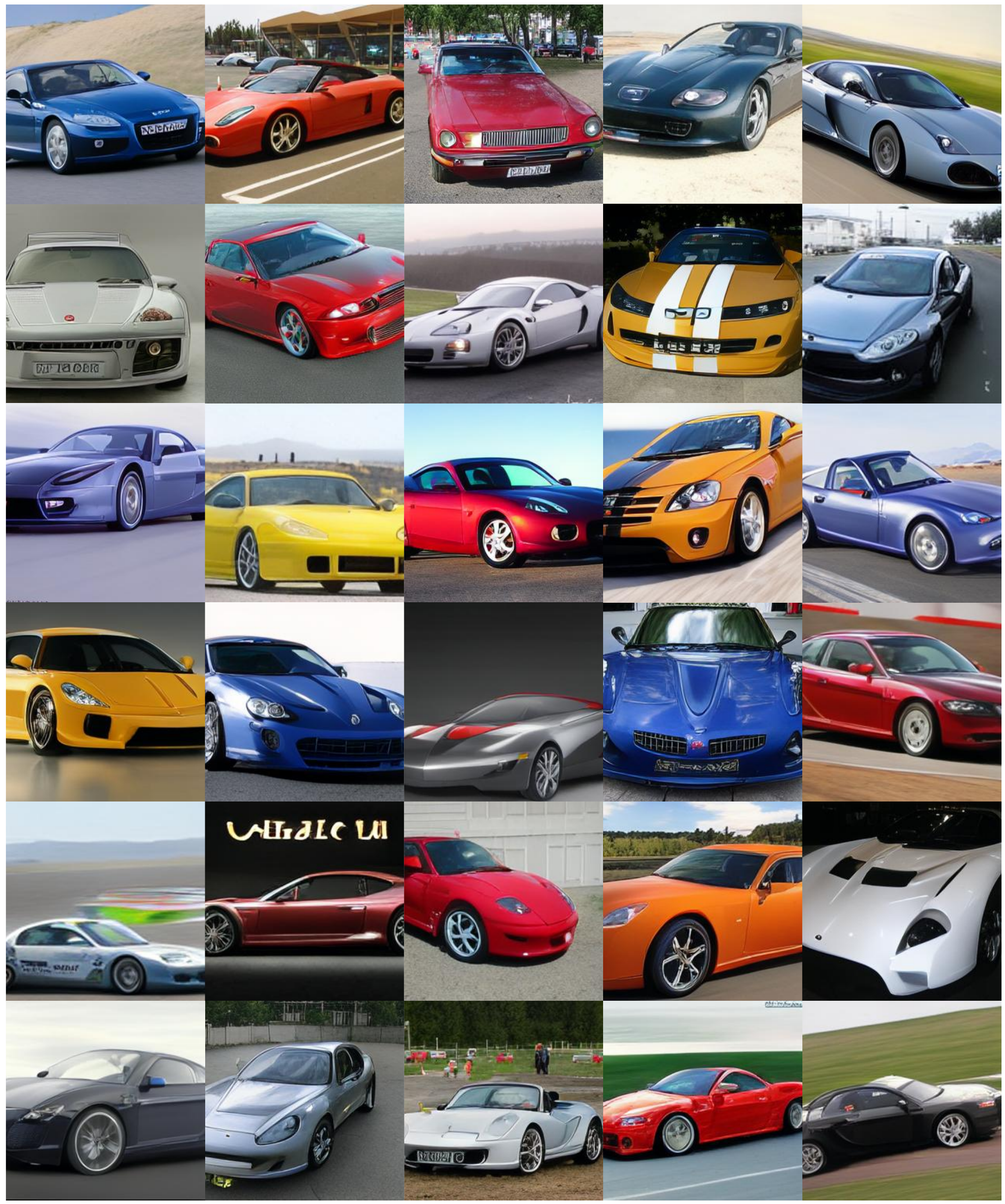}
    \caption{   256×256 LlamaGen-XL + \modelunidir{} samples. Classifier-free guidance scale = 1.75. Class label = ``sport car'' (817).   } 
    \label{fig:car E}
\end{figure}

\begin{figure}[ht]
  \centering
     \includegraphics[width=1\linewidth]{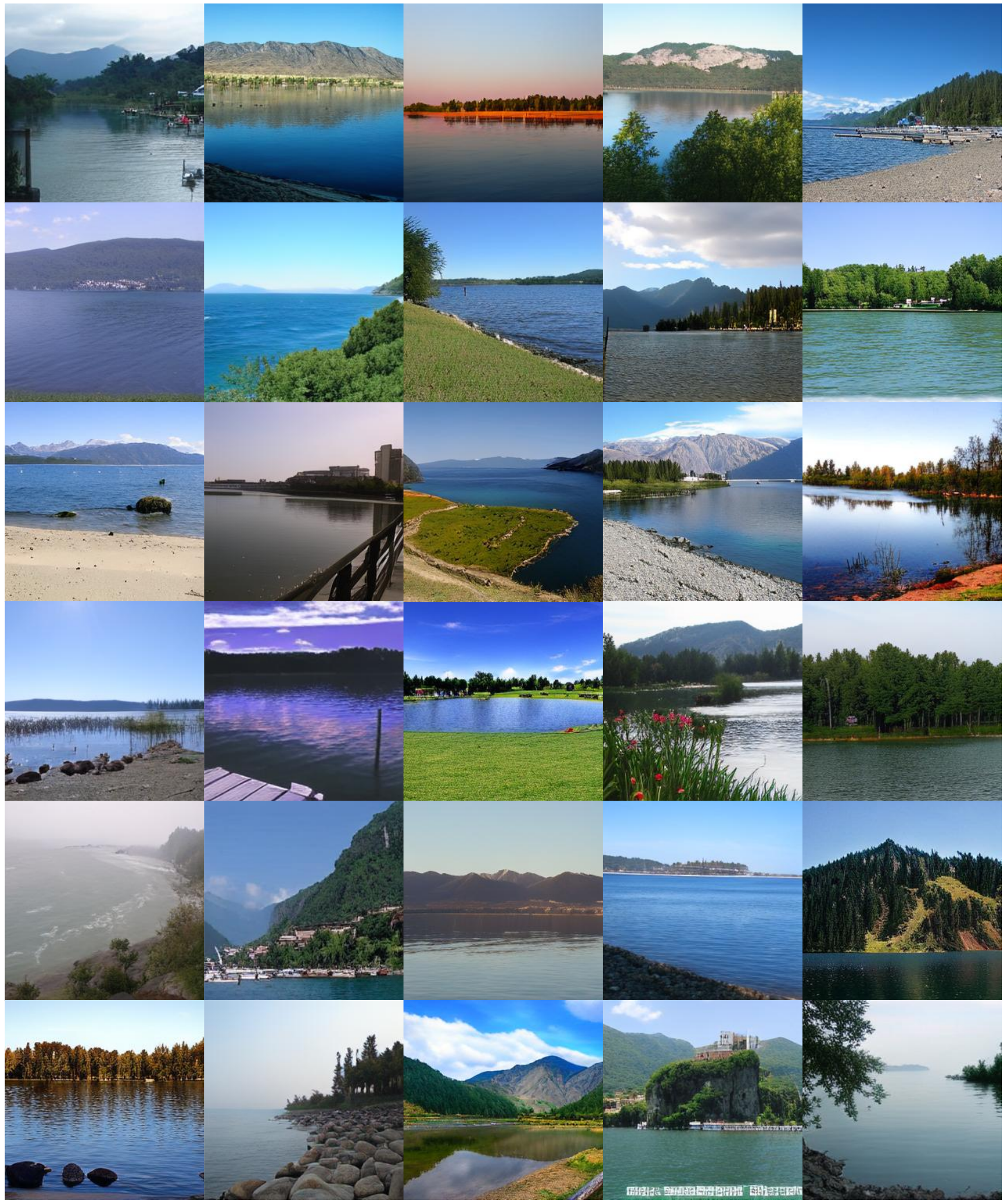}
    \caption{   256×256 LlamaGen-XL + \modelbidir{} samples. Classifier-free guidance scale = 1.75. Class label = ``lake shore'' (975).    } 
    \label{fig:lake I}
\end{figure}

\begin{figure}[ht]
  \centering

     \includegraphics[width=1\linewidth]{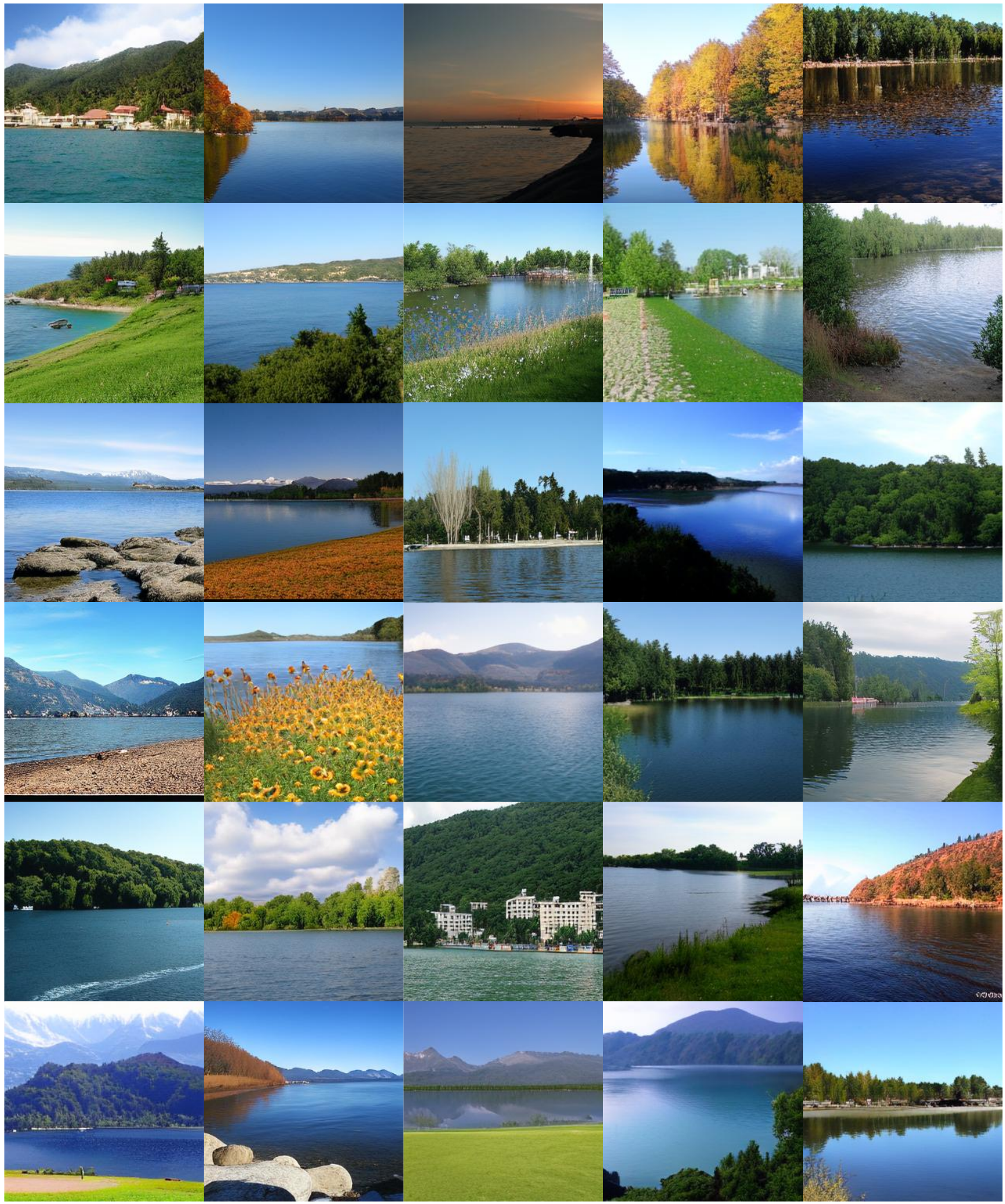}
    \caption{   256×256 LlamaGen-XL + \modelunidir{} samples. Classifier-free guidance scale = 1.75. Class label = ``lake shore'' (975).  } 
    \label{fig:lake E}
\end{figure}

Below, we show additional uncurated generation results on ImageNet 256×256 from the LlamaGen-XL + \modelname{} in \cref{fig:dog I}, \cref{fig:dog E}, \cref{fig:car I}, \cref{fig:car E}, \cref{fig:lake I} and \cref{fig:lake E}. We use classifier-free guidance with scale 1.75.


   \vspace{106mm}

\end{document}